\documentclass[%
nofootinbib,
 amsmath,amssymb,
 aps,
 notitlepage,
 onecolumn,
floatfix,
]{revtex4-2}
\usepackage{natbib}
\usepackage{graphicx} % Required for inserting images
\usepackage{blindtext}
\usepackage{geometry}
\usepackage[section]{placeins}
\usepackage{amsfonts}
\usepackage{amssymb}
\usepackage{hyperref}
\hypersetup{colorlinks=true,allcolors=blue}
\usepackage[mathlines]{lineno}% Enable numbering of text and display 
\usepackage{mathtools}
\usepackage{nccmath}
\usepackage{float}
\usepackage{appendix}
\usepackage{listings}

\usepackage{xcolor}

\definecolor{codegreen}{rgb}{0,0.6,0}
\definecolor{codegray}{rgb}{0.5,0.5,0.5}
\definecolor{codepurple}{rgb}{0.58,0,0.82}
\definecolor{backcolour}{rgb}{0.95,0.95,0.92}
\usepackage{inconsolata}
\lstdefinestyle{mystyle}{
    backgroundcolor=\color{backcolour},   
    commentstyle=\color{codegreen},
    keywordstyle=\color{magenta},
    numberstyle=\tiny\color{codegray},
    stringstyle=\color{codepurple},
    basicstyle=\ttfamily\small,
    breakatwhitespace=false,         
    breaklines=true,                 
    captionpos=b,                    
    keepspaces=true,                 
    showstringspaces=false,
    tabsize=2
}
\usepackage{xpatch}
\usepackage{realboxes}
\makeatletter
\xpretocmd\lstinline{\Colorbox{backcolour}\bgroup\appto\lst@DeInit{\egroup}}{}{}
\makeatother
\usepackage[T1]{fontenc}
\usepackage[mathletters]{ucs}
\usepackage[utf8]{inputenc}

\usepackage{caption}
\usepackage{dcolumn}% Align table columns on decimal point
\usepackage{bm}% bold math
\usepackage{orcidlink}
\usepackage[utf8]{inputenc}
\newcommand{\code}[1]{{\Colorbox{backcolour}{\small\texttt{#1}}}}

\usepackage{orcidlink}
\usepackage{appendix}
\usepackage{caption}
\begin{document}
\title{An introduction to graphical tensor notation for mechanistic interpretability}
\author{Jordan K. Taylor\,\orcidlink{0000-0002-5799-0557}}
\email[]{jordantensor@gmail.com}
 % \altaffiliation[Also at ]{Physics Department, XYZ University.}%Lines break automatically or can be forced with \\
\affiliation{%
 School of Mathematics and Physics, University of Queensland, Brisbane, Queensland 4072, Australia
}%
\maketitle
Graphical tensor notation is a simple way of denoting linear operations on tensors, originating from physics. Modern deep learning consists almost entirely of operations on or between tensors, so easily understanding tensor operations is quite important for understanding these systems. This is especially true when attempting to reverse-engineer the algorithms learned by a neural network in order to understand its behavior: a field known as mechanistic interpretability. It's often easy to get confused about which operations are happening between tensors and lose sight of the overall structure, but graphical notation\footnote{Graphical tensor notation is also referred to as \href{https://tensornetwork.org/diagrams/}{tensor-network notation}~\cite{stoudenmire2022tensornetworkorg}, \href{http://homepages.math.uic.edu/~kauffman/Penrose.pdf\#page=5}{Penrose graphical notation}~\cite{penrose1971applications}, \href{https://en.wikipedia.org/wiki/ZX-calculus}{ZX-calculus}, or \href{https://ncatlab.org/nlab/show/string+diagram}{string-diagram notation} depending on the context.} makes it easier to parse things at a glance and see interesting equivalences.
\begin{figure}[h]
    \centering
    \includegraphics[width=0.9\linewidth]{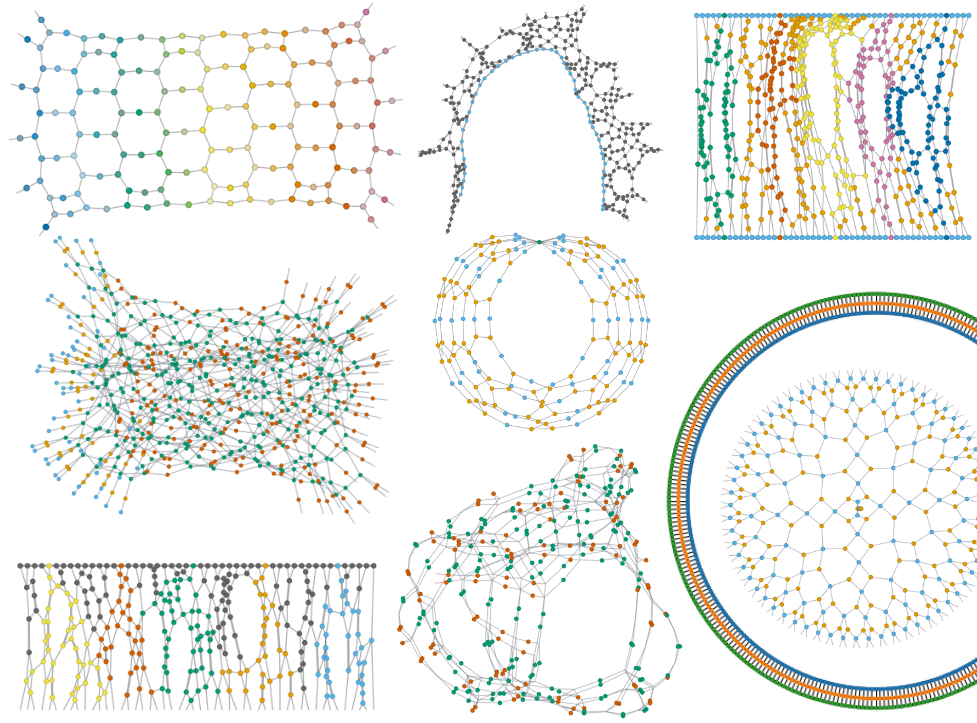}
    \caption{Some examples of graphical tensor notation from the \href{https://quimb.readthedocs.io/en/latest/index.html}{QUIMB python package}~\cite{gray2018quimb}}
\end{figure}

The first half of this document introduces the notation and applies it to some decompositions (SVD, CP, Tucker, and tensor-network decompositions), while the second half applies it to some existing some foundational approaches for mechanistically understanding language models, loosely following \href{https://transformer-circuits.pub/2021/framework/index.html}{A Mathematical Framework for Transformer Circuits}~\cite{elhage2021mathematical}, then constructing an example ``induction head'' circuit in graphical tensor notation. 

My explanations of graphical tensor notation are partly based on existing explanations such as in \href{https://www.math3ma.com/blog/matrices-as-tensor-network-diagrams}{the math3ma blog}~\cite{bradley2021matrices}, \href{https://simonverret.github.io/2019/02/16/tensor-network-diagrams-of-typical-neural-network.html}{Simon Verret's blog}~\cite{verret2019tensor}, \href{https://tensornetwork.org/diagrams/}{tensornetwork.org}~\cite{stoudenmire2022tensornetworkorg}, \href{https://www.tensors.net/p-tutorial-1}{tensors.net}~\cite{evenbly2023tensorsnet}, \href{https://arxiv.org/abs/1603.03039}{Hand-waving and Interpretive Dance: An Introductory Course on Tensor Networks}~\cite{bridgeman2017hand}, and \href{https://arxiv.org/abs/2303.13565}{An Intuitive Framework for Neural Learning Systems}~\cite{xu2023graph}. Feel free to skip around and look at interesting diagrams rather than reading this document start to finish.

\tableofcontents
\newpage
% \sectioninfo{}
\section{Tensors}
Practically, tensors in our context can just be treated as arrays of numbers.\footnote{Technically tensors are abstract multilinear maps, rather than just arrays of numbers. However the two are equivalent once a basis for the multilinear space has been chosen.} In graphical notation (first \href{http://homepages.math.uic.edu/~kauffman/Penrose.pdf\#page=5}{introduced by Roger Penrose in 1971}~\cite{penrose1971applications}), tensors are represented as shapes with ``legs'' sticking out of them. A vector can be represented as a shape with one leg, a matrix can be represented as a shape with two legs, and so on. I'll also represent everything in PyTorch code for clarity.

\[
    \includegraphics[width=0.73\linewidth]{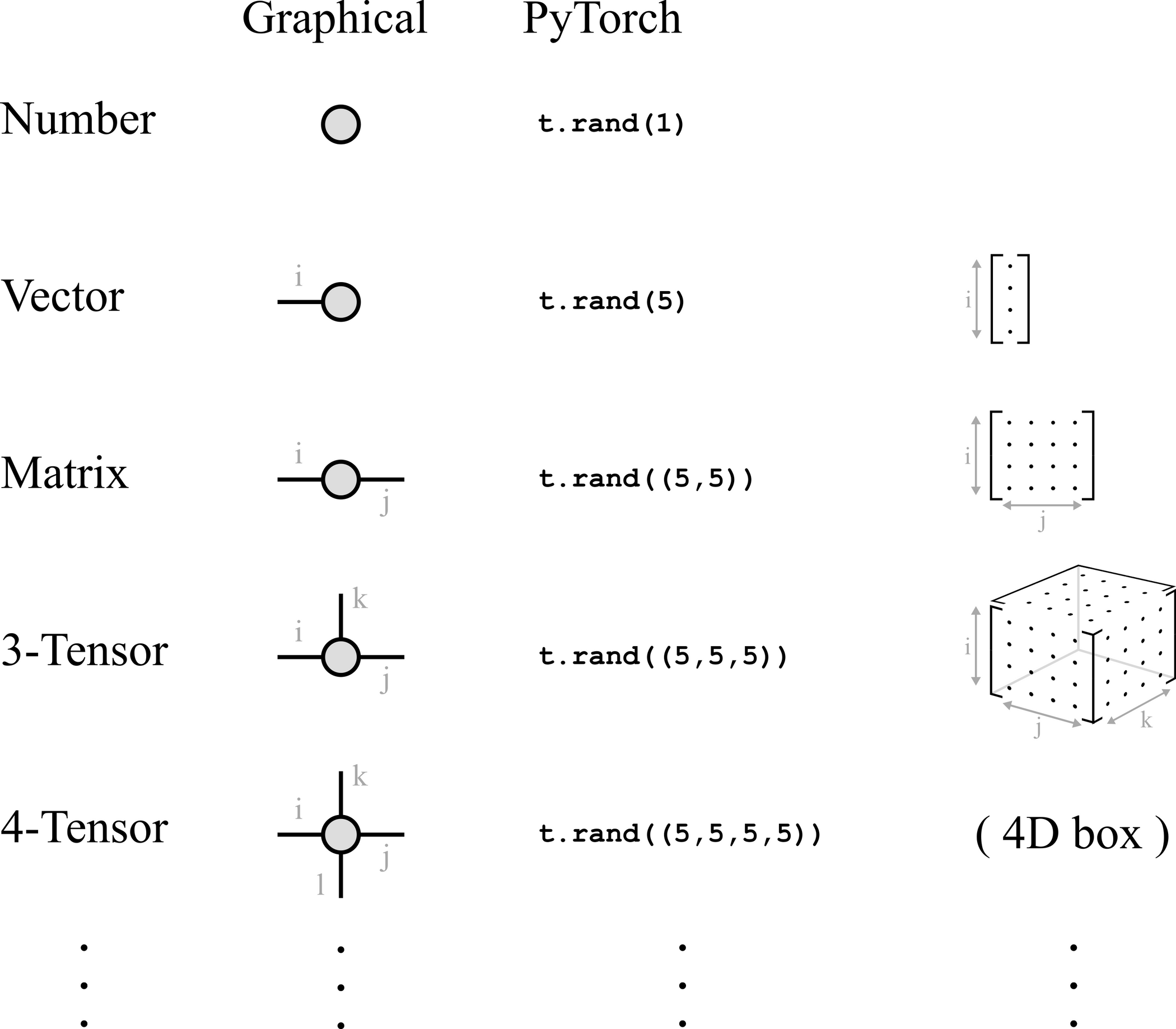}
\]

Each leg corresponds to an index of the tensor - specifying an integer value for each leg of the tensor addresses a number inside of it:

\[
    \includegraphics[width=0.55\linewidth]{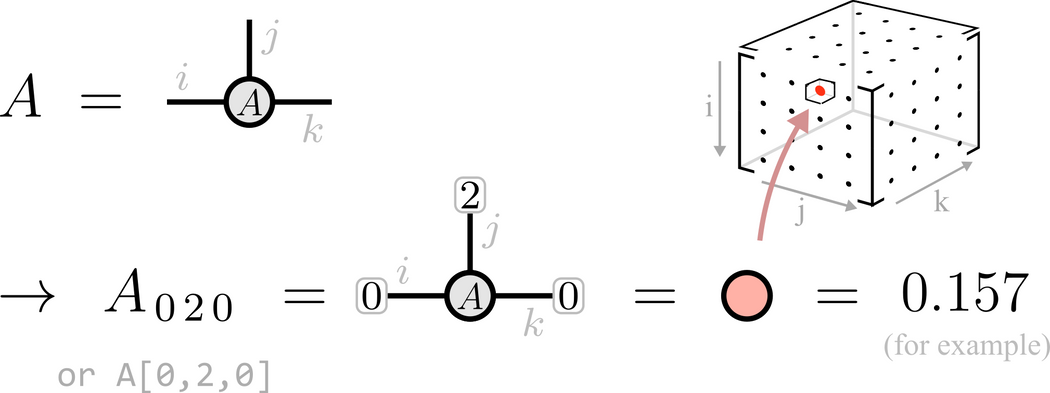}\;,
\]
where $0.157 $ happens to be the number in the $(i=0,\,j=2,\,k=0) $ position of the tensor $A $. In python code, this would be \code{A[0,2,0]}. 

The amount of memory required to store a tensor grows exponentially with the number of legs,\footnote{A tensor with $N $ legs each of dimension $d $ contains $d^N $ numbers.} so tensors with lots of legs are usually represented only implicitly: decomposed as operations between many smaller tensors.

\newpage
\section{Operations}
% \sectioninfo{A very long text, a very long text, a very long text, a very long text, a very long text, a very long text, a very long text, a very long text}
The notation only really becomes useful when things get more complicated, but let's start as simple as possible. Multiplying two numbers together (\code{y = a * b}) in graphical tensor notation just involves drawing them nearby:
\[
    \includegraphics[width=0.6\linewidth]{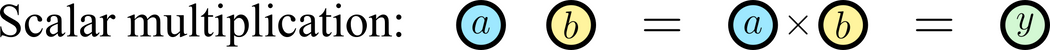} \;.
\]

The next easiest thing to represent in this notation is a bit more obscure: the outer product between two vectors (\code{t.outer(a, b)} or \code{einsum(a, b, `i, j, -> i j')}). Known more generally as a tensor product, this operation forms a matrix out of the vectors, where each element in the matrix is a product of two numbers: $Y_{i\,j}=a_ib_j $ or \code{Y[i,j] = a[i] * b[j]} (for example \code{Y[0,0] = a[0] * b[0]},  \code{Y[1,0] = a[1] * b[0]} and so on). Simply drawing two tensors nearby implies a tensor product:
\[
    \includegraphics[width=0.6\linewidth]{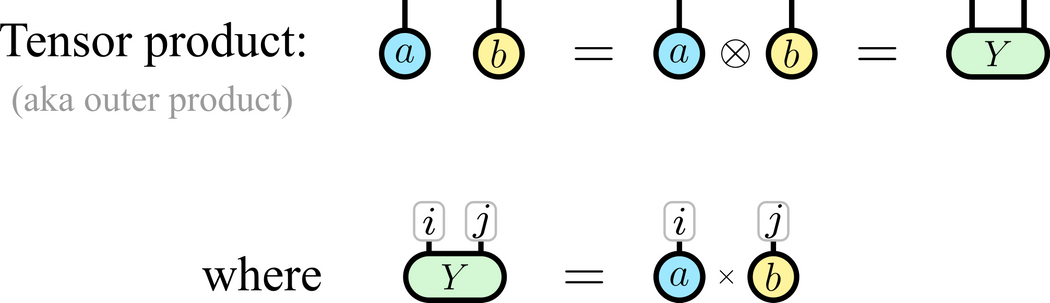} \;.
\]

The next simplest example will probably be more familiar: the dot product between two vectors: \code{t.dot(a,b)} or \code{a @ b} or \code{einsum(a, b, `i, i, -> ')}, which can be represented by connecting the legs of two vectors:
\[
    \includegraphics[width=0.37\linewidth]{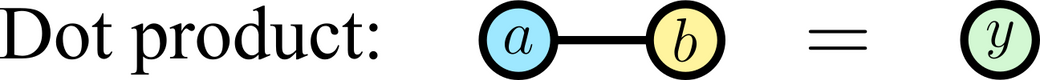} \;.
\]
Connected legs like this indicate that two tensors share the same index, and a summation is taken over that index. Here the result is a single number, formed from a sum of products: $y = \sum_i a_ib_i $ or \code{y = a[0] * b[0]  +  a[1] * b[1]  +  a[2] * b[2] + ...}

Connecting legs like this is known more generally as tensor contraction or Einstein summation.  Let's take a look at all of the most common kinds of contractions between vectors and matrices:
\[
    \includegraphics[width=\linewidth]{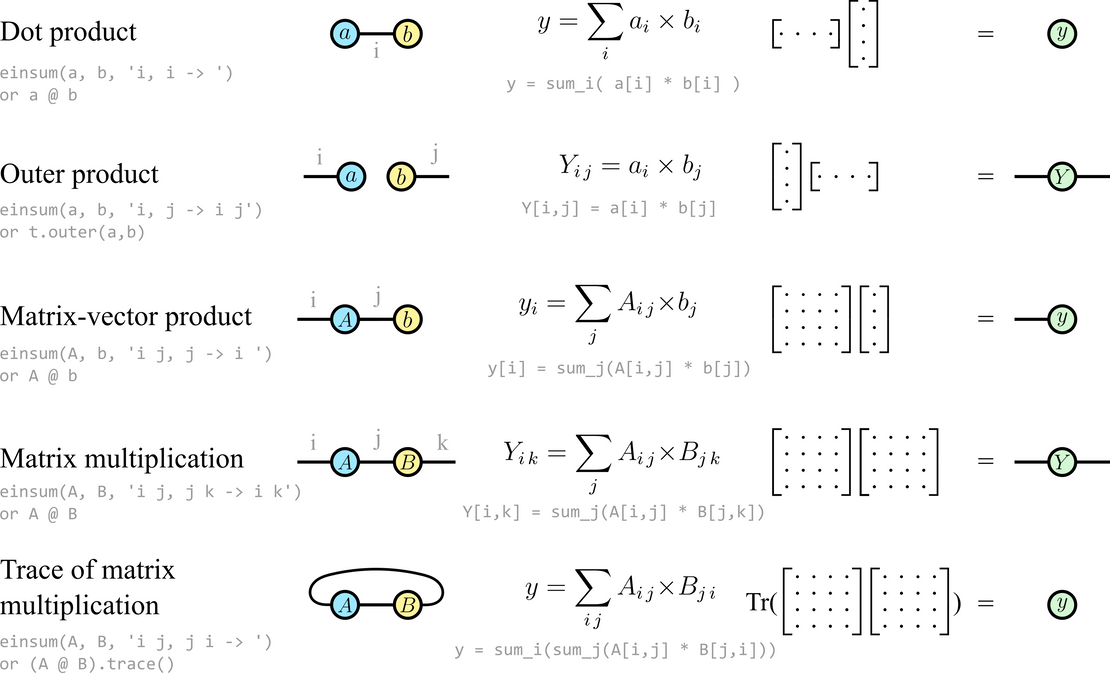} \;.
\]
In every case you can tell how many legs the resulting tensor will have by the number of uncontracted ``free'' legs on the left. \newpage

We can also represent single-tensor operations, such as the transpose of a matrix:
\[
\begin{tabular}{ c  c }
Graphical notation & einops / PyTorch \\
\hline\rule{0pt}{4.5ex}   
 $\vcenter{\hbox{\includegraphics[height=24pt]{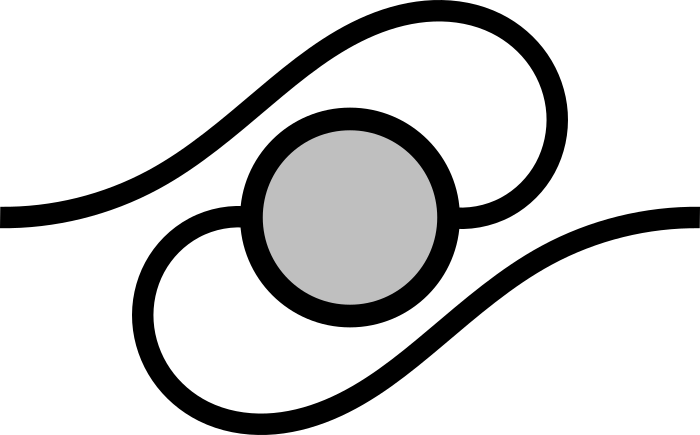}}}$ & \code{rearrange(A, `i j -> j i')} or \code{A.transpose(0, 1)} \\
\end{tabular}
\]
the rearranging of tensor indices:
\[
\begin{tabular}{ c  c }
Graphical notation & einops / PyTorch \\
\hline\rule{0pt}{4.5ex}   
 $\vcenter{\hbox{\includegraphics[height=45pt]{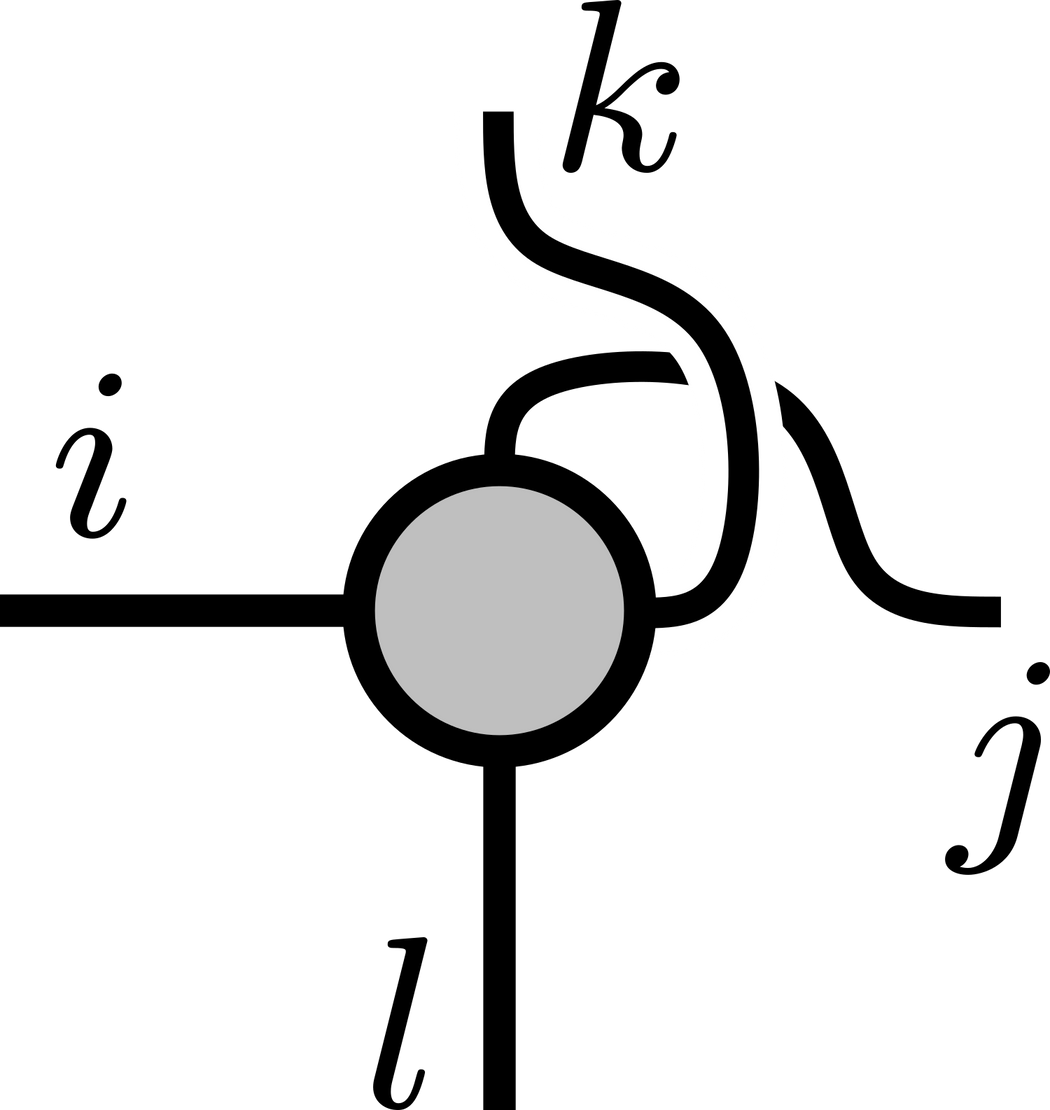}}}$ & \code{rearrange(T, `i j k l -> i k j l')} or \code{T.transpose(1, 2)} \\
\end{tabular}
\]
and the reshaping (flattening) of a tensor into a matrix by grouping some of its legs:
\[
\begin{tabular}{ c  c }
Graphical notation & einops / PyTorch \\
\hline\rule{0pt}{4.5ex}   
 $\vcenter{\hbox{\includegraphics[height=21pt]{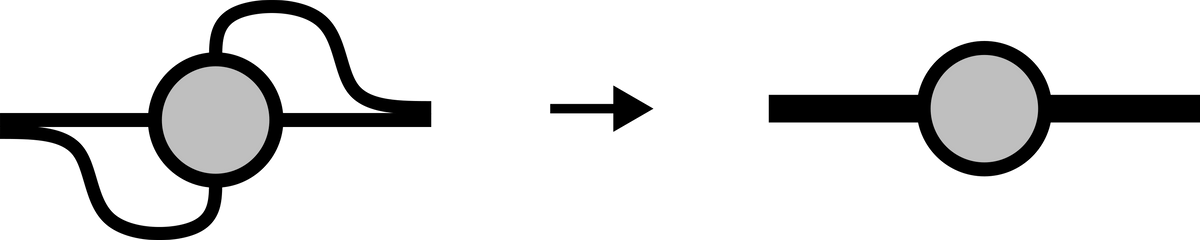}}}$ &  ~\code{rearrange(T, `i j k l -> (i l) (k j)')}  \\
\end{tabular}
\]
where thicker lines are used to represent legs with a larger dimension. Of course you can also split legs rather than grouping them:
\[
\begin{tabular}{ c  c }
Graphical notation & einops / PyTorch \\
\hline\rule{0pt}{4.5ex}   
 $\vcenter{\hbox{\includegraphics[height=15pt]{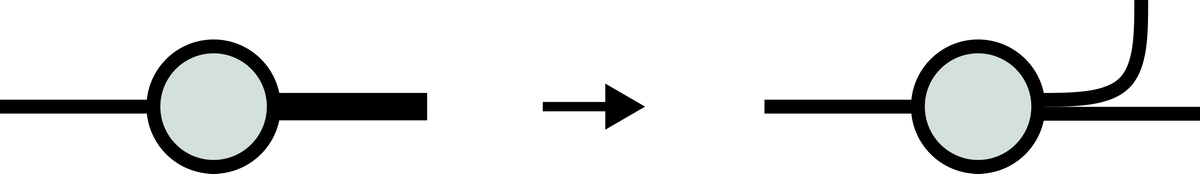}}}$ &   ~\code{rearrange(M, `i (j k) -> i j k', j=int(sqrt(M.shape[-1]))))}  \\
\end{tabular}
\]

Various relationships also become intuitive in graphical notation, such as the cyclic property of the trace  $\mathrm{Tr}(AB)=\mathrm{Tr}(BA) $:
\[
    \includegraphics[width=0.5\linewidth]{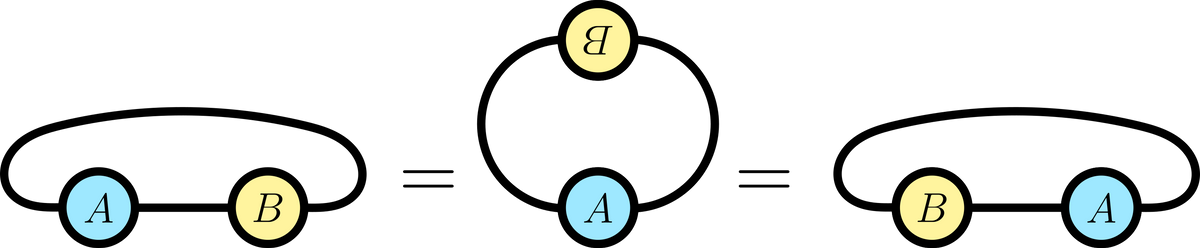} \;,
\]
or if you prefer transposes rather than upside-down tensors:
\[
    \includegraphics[width=0.8\linewidth]{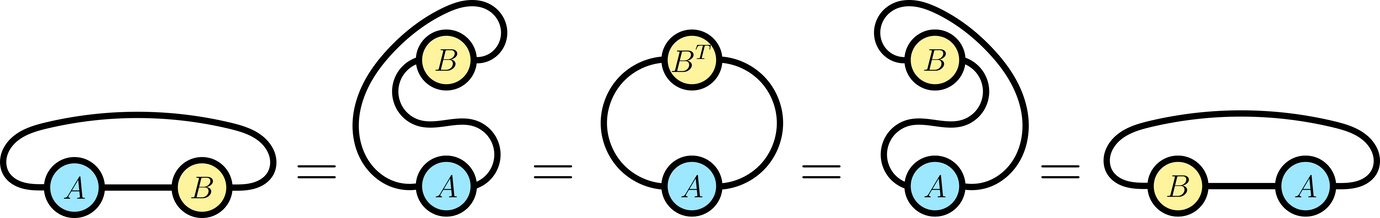} \;.
\]

But graphical notation is most useful for representing unfamiliar operations between many tensors. One example in this direction is  $\sum_{\alpha\beta} A_{i\alpha\beta} \, v_{\beta} \, B_{\alpha \beta j} = M_{ij}, $ which can be represented in graphical notation as
\[
    \includegraphics[width=0.9\linewidth]{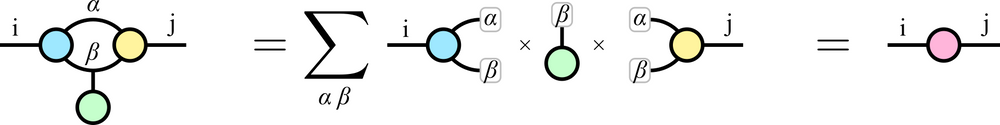} \;,
\]

or in einops as \code{M = einsum(A,v,B,`i $\alpha$ $\beta$, $\beta$, $\beta$ $\alpha$ j -> i j')}. The middle part of the graphical notation here shows that the number in each $i$, $j$ position of the final matrix can be calculated with a sum over every possible indexing of the internal legs $\alpha$ and $\beta$, 

where each term in the sum consists of three numbers being multiplied (though in practice the contraction should be calculated in a much more efficient way). 

Graphical notation really comes into its own when dealing with larger networks of tensors. For example, consider the contraction 
\[\label{eq:two_leg_ladder_old_notation}
    \sum_{i\,j\,k\,l\,m\,n\,o\,p\,q\,r\,s\,t\,u} A_{ij} V_{ir} B_{jkl} W_{rks} C_{lmn} X_{smt}  D_{nop} Y_{tou}  E_{pq} Z_{uq},
\]
which is tedious to parse: indices must be matched up across tensors, and it is not immediately clear what kind of tensor (eg. number, vector, matrix ...) the result will be.
Needless to say, the einsum code is about as bad: \\\code{einsum(A,V,B,W,C,X,D,Y,E,Z,`i j, i r, j k l, r k s, l m n, s m t, n o p, t o u, p q, u q -> ')}. But in graphical notation this is 
\[
    \includegraphics[width=0.27\linewidth]{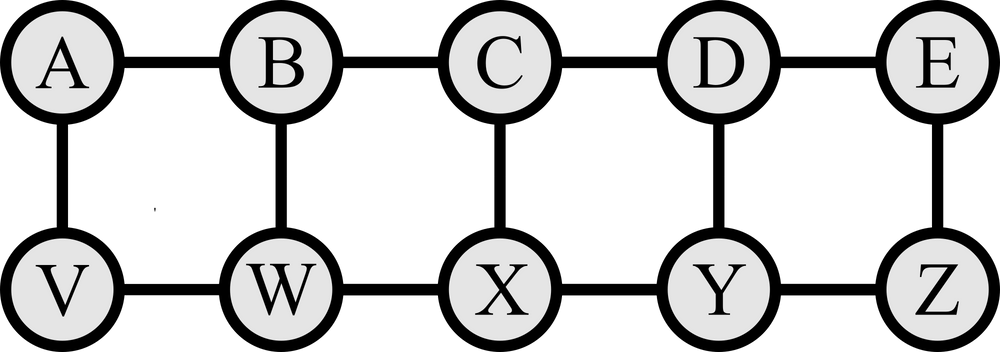} \;,
\]
and we can immediately see which tensors are to be contracted, and that the result will be a single number. Contractions like this can be performed in any order. Some ways are much more efficient than others,\footnote{Consider contracting along the top line first:
\[\includegraphics[width=0.85\linewidth]{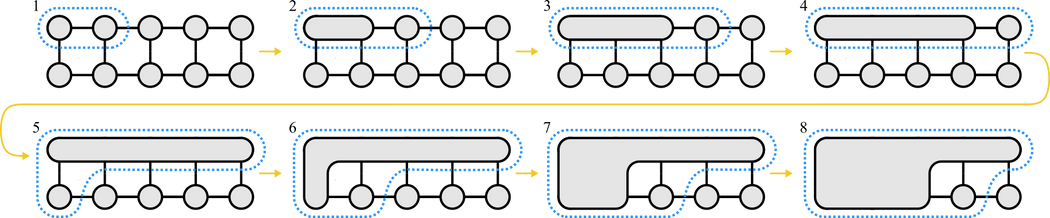}\]
this has a cost exponential in the number of tensors, because an intermediate tensor is created with $N/2 $ legs. A much more efficient order is
\[\includegraphics[width=0.85\linewidth]{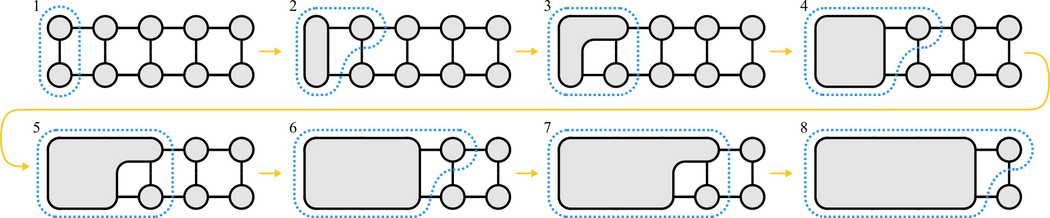}\]
which limits the the intermediate tensors to no more than three legs, and scales linearly with the number of tensors. 
In general, finding the optimal order in which to contract a tensor network \href{https://www.worldscientific.com/doi/abs/10.1142/S0129626497000176}{is an NP-hard problem}~\cite{chi1997optimizing}, let alone actually performing the contraction, \href{https://link.springer.com/article/10.1007/s00037-000-0170-4}{which is \#P-hard in general}~\cite{damm2002complexity_tensor_calculus}. Usually though, fairly simple contraction order heuristics and approximation techniques get you relatively far. } but they all get the same answer eventually.

Tensor networks (einsums) like this also have a nice property that, if the tensors are independent (not copies or functions of each other), then a derivative of the final result with respect to one of the tensors can be calculated just by ``poking a hole'' and removing that tensor:
\[
    \includegraphics[width=0.7\linewidth]{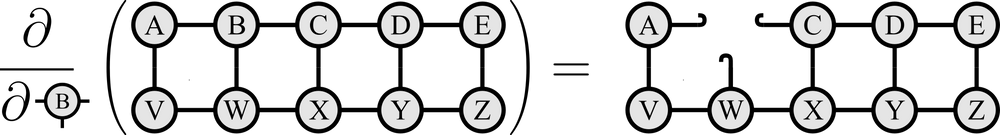} \;.
\]
This is because einsums are entirely linear (or multilinear, at least).

\section{Special kinds of tensors}

Different kinds of tensors are often drawn using different shapes. Firstly, it's common to represent an identity matrix as a single line with no shape in the middle:
\[
    \includegraphics[width=0.65\linewidth]{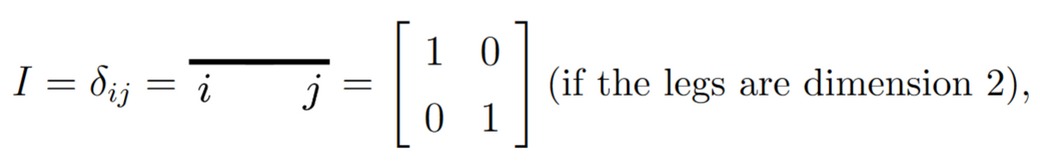} 
\]
(or \code{t.eye(2)}) where we have used the delta notation $\delta_{ij} $ because the elements of the identity matrix are equivalent to the Kronecker delta of the indices: $1 $ if $i=j $ and zero otherwise. You can also extend this notation to the three-leg delta tensor, which has ones only along the $i=j=k $ diagonal:
\[
    \includegraphics[width=0.65\linewidth]{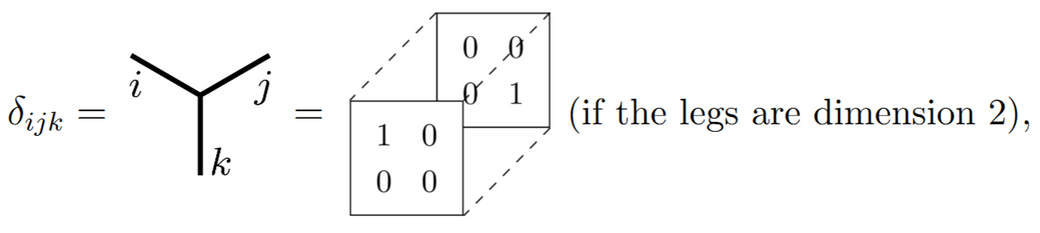} 
\]
and so on for delta tensors with more legs. Among other things, this lets us represent diagonal matrices using vectors:
\[
    \includegraphics[width=0.34\linewidth]{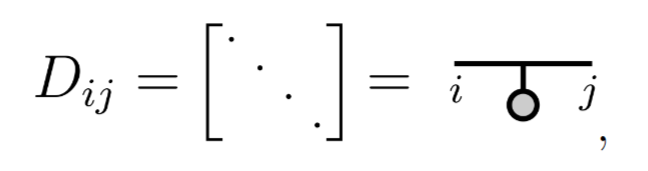}
\]
where the vector in grey contains just the elements on the diagonal. Typically though, it would be inefficient to actually code a contraction with a delta tensor made of actual numbers: it's much faster to just rearrange or reindex the relevant data directly. Still, whenever you see any line in a tensor network diagram, you can imagine a delta tensor implicitly sitting there. 

Triangles are often used to represent isometric matrices: linear maps which preserve the lengths of vectors (eg. performing a rotation), even if they might embed these vectors into a larger-dimensional space.  A matrix $V $ is isometric if it can be contracted with its own (conjugate) transpose to yield the identity matrix. Graphically, 
\[
\begin{tabular}{ c c c }
Graphical notation & Math & einops / PyTorch \\
\hline\rule{0pt}{4.5ex}   
 $\vcenter{\hbox{\includegraphics[height=20pt]{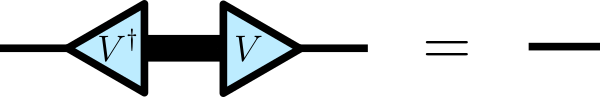}}}$ &   ~ $V^\dagger V = I $ & ~\code{t.transpose(t.conj(V)) @ V == t.eye(V.shape[-1])}  \\
\end{tabular}
\]
where the tip of the triangle points towards the smaller dimension. However the reverse is not true when the matrix is not square, because some vectors will inevitably get squashed when mapping from high to low dimensions:
\[
    \includegraphics[height=20pt]{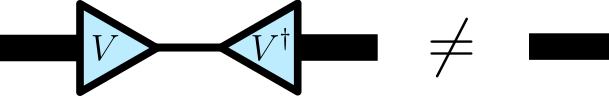} \;.
\]
Square isometries are known as orthogonal matrices (or unitary matrices if they contain complex numbers), and are often represented with squares or rectangles:
\[
    \includegraphics[height=15pt]{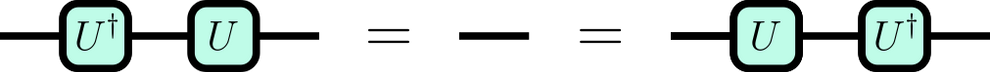} \;.
\]
When isometries have more than just two legs, their legs can be grouped by whether they go into the edge or the tip of the triangle, and similar relationships hold:
\[
    \includegraphics[width=110pt]{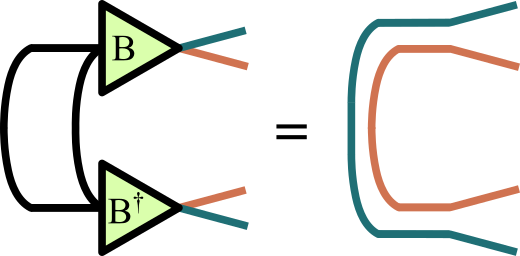} \;.
\]
Finally, here's a silly looking related graphical equation:
\[
    \includegraphics[height=15pt]{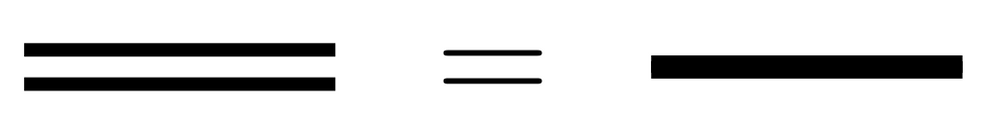} \;.
\]
It says that the flattened tensor product (Kronecker product) of two identity matrices is another identity matrix. In pytorch / einops, this is \code{t.kron(t.eye(5), t.eye(3)) == t.eye(5*3)} or \code{einsum(t.eye(5), t.eye(3), `i j, k l -> (i k) (j l)') == t.eye(5*3)}.\footnote{ technically  \code{rearrange(einsum(t.eye(5), t.eye(3), `i j, k l -> i k j l'), `i k j l -> (i k) (j l)')} as shape rearrangement isn't yet supported within an einsum call.}

\newpage
\section{Decompositions (SVD, CP, Tucker)}
\vspace{-12pt}
\textit{[Feel free to skip to section \ref{Neural networks} on neural networks]}\\

The Singular Value Decomposition (SVD) allows any matrix $M $ to be decomposed as $M=UDV^\dagger $, where $U $ and $V $ are isometric matrices, and $D $ is a diagonal matrix:
\[
    \includegraphics[height=25pt]{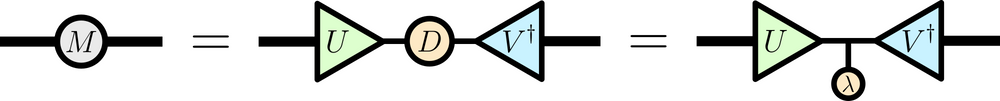} \;,
\]
where $\lambda $ is the vector of non-negative \textit{singular values} making up the diagonal elements of $D $. Or in torch / einops: \code{U, $\lambda$, V = t.svd(M)} and \code{allclose(M, einsum(U, $\lambda$, t.conj(V), `i j, j, k j -> i k'))}. 

There are many intuitive ways of thinking about the SVD.\footnote{See  \href{https://www.lesswrong.com/posts/iupCxk3ddiJBAJkts/six-and-a-half-intuitions-for-svd}{Six (and a half) intuitions for SVD}~\cite{callummcdougall2023six}, for example.} Geometrically, the SVD is often thought of as decomposing the linear transformation $M $ into a ``rotation'' $V^\dagger $, followed by a scaling $D $ of the new basis vectors in this rotated basis, followed by another ``rotation'' $U $. However in the general case where $U $ and $V^\dagger $ are complex-valued isometries rather than just rotation matrices, this geometric picture becomes harder to visualize.

Instead, it is also useful to think of the SVD as sum of outer products of vectors:
\[
    \includegraphics[height=37pt]{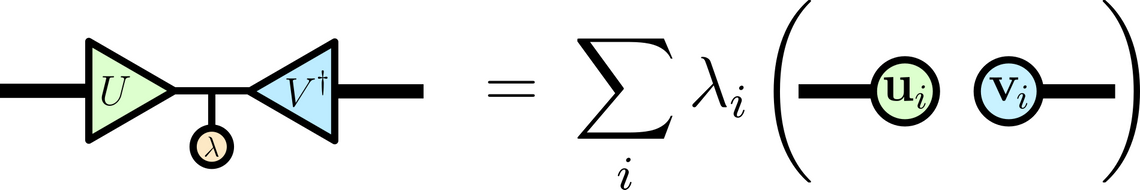} \;,
\]
where $\textbf{u}_1,\, \textbf{u}_2,\, \dots $ are the orthonormal vectors from the columns of $U $, and $\textbf{v}^\dagger_1,\, \textbf{v}^\dagger_2,\, \dots $ are the orthonormal vectors from the rows of $V^\dagger $. 

\begin{figure}
    \centering
    \includegraphics[width=0.75\linewidth]{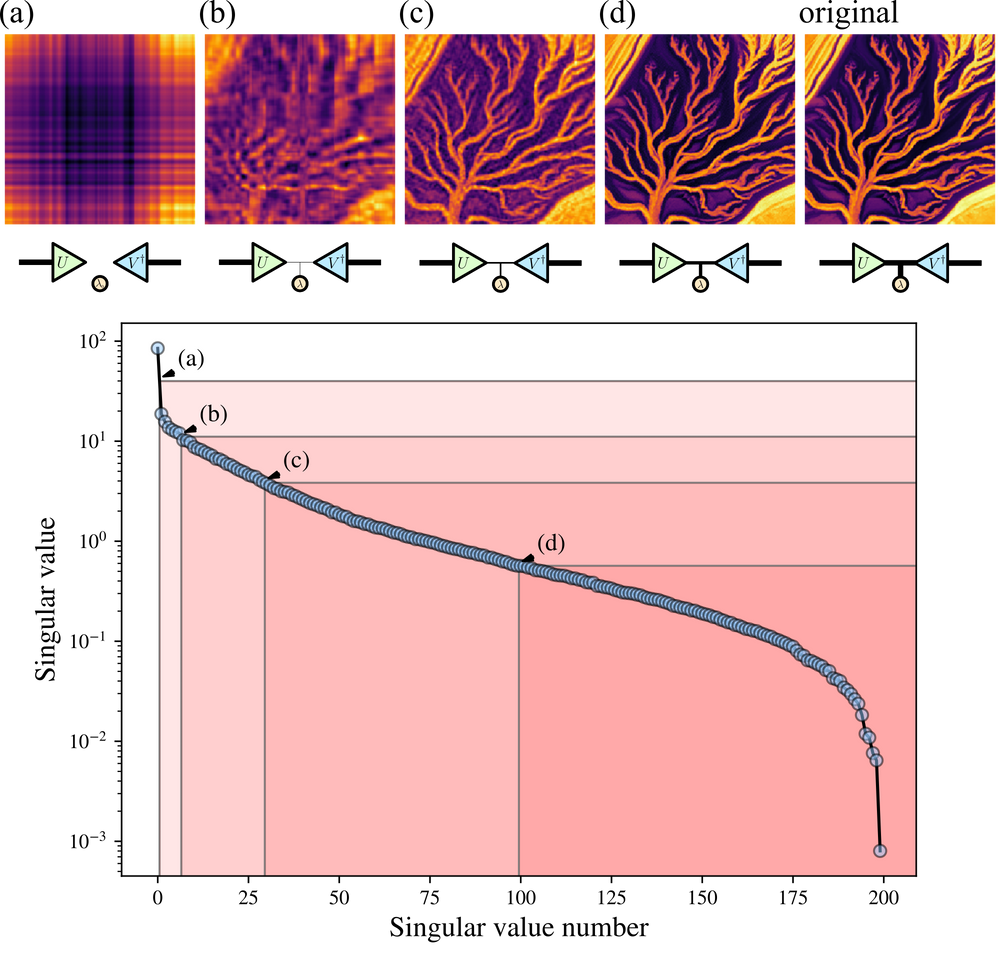}
    \caption{A matrix can be compressed by performing a singular value decomposition and discarding the smallest singular values. Here I treat an image as a matrix, and perform various levels of truncation, with the discarded singular values shown in the red shaded regions of the plot. (a) shows just one singular value kept: the matrix is approximated as a single outer-product of two vectors, scaled by the first singular value. (b) shows 7 singular values, (c) 30, and (d) 100 kept out of the 200 singular values in the full decomposition.}
    \label{fig:svd_trunc}
\end{figure}

The size of each singular value $\lambda_i $ corresponds to the importance of each corresponding outer product. The number of nonzero singular values is known as the ``rank'' of the matrix $M $. When some singular values are sufficiently close to zero, their terms can be omitted from the sum, lowering the rank of $M $. The effects of this low-rank approximation can be seen by treating an image as a matrix, and compressing it by performing an SVD and discarding the small singular values, as shown in figure \ref{fig:svd_trunc}.\footnote{This is not the most natural way of representing a matrix or the effects of an SVD, because images have a notion of locality between nearby pixels, whereas nearby entries in a matrix are treated as unrelated. Still, it's intuitive and easy to visualize. See part \href{https://www.lesswrong.com/posts/iupCxk3ddiJBAJkts/six-and-a-half-intuitions-for-svd\#6B__Information_Compression}{6B of Six and a half intuitions}~\cite{callummcdougall2023six} for a more natural SVD compression example. }

In fact, performing an SVD and keeping only the largest $k $ singular values $\lambda_1 \cdots \lambda_k $provides the best possible rank-$k $ approximation of the original matrix $M $. This is known as the Eckart–Young theorem, and is true regardless of whether the ``best'' approximation is defined by the spectral norm, the Frobenius norm, or any other unitarily invariant norm~\cite{schmidt1907theorie_eckart_young, eckart1936approximation_eckart_young, mirsky1960symmetric_eckart_young}. The error in this approximation is determined by the total weight of the singular values thrown away.

General tensors can also be decomposed with the SVD by grouping their legs, forming a bipartition of the tensor:
\[
    \includegraphics[height=28pt]{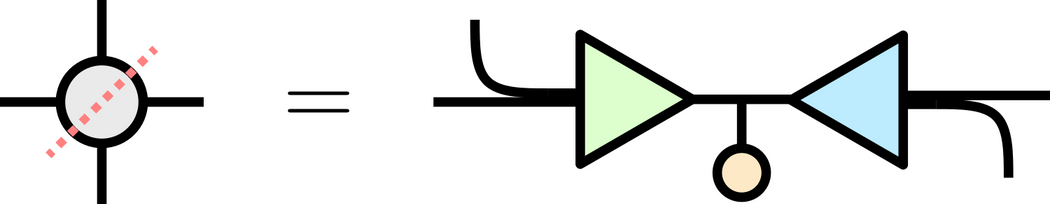} \;.
\]

The SVD also has some higher-order generalizations, such as the CP and Tucker decompositions. These decompositions work directly on tensors with any number of legs, without requiring that legs be grouped into an effective matrix. The simplest generalization of the SVD is the CP (Canonical Polyadic or CANDECOMP/PARAFAC decomposition~\cite{hitchcock1927expression_cp_decomposition, carroll1970analysis_cp_decomposition, harshman1970foundations_cp_decomposition}, which extends the SVD pattern naturally to more legs:
\[
    \includegraphics[height=45pt]{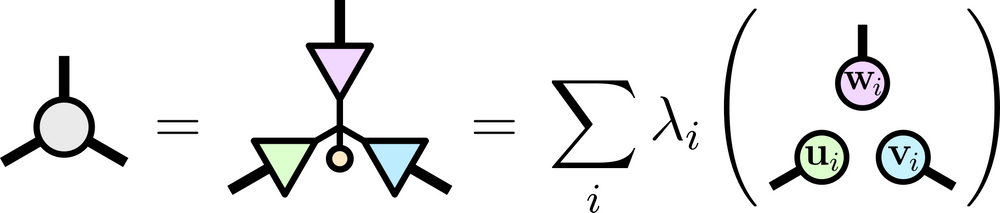} \;,
\]
or in code:
\begin{lstlisting}[language=Python]
T = t.rand((2,3,4))
s, [U, V, W] = tensorly.decomposition.parafac(T, rank=9, tol=1e-12)
O = einsum(s, U, V, W, 's, i s, j s, k s -> i j k')
t.allclose(T, O, rtol=1e-3)
\end{lstlisting}
Whereas the Tucker decomposition is a relaxation of the CP decomposition where the core tensor is not restricted to be diagonal:~\cite{tucker1966some_mathematical_notes_decomposition}
\[
    \includegraphics[height=45pt]{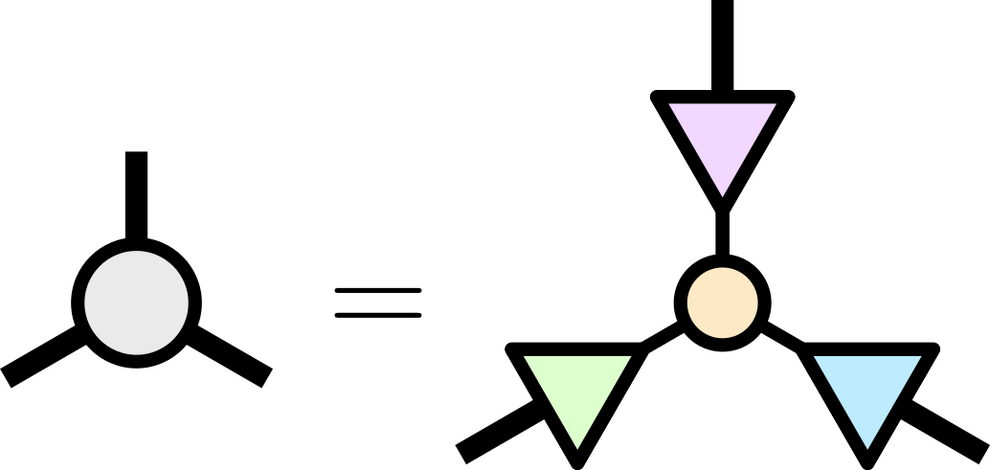} \;,
\]
or in code:
\begin{lstlisting}[language=Python]
T = t.rand((10,10,10))
C, [U, V, W] = tensorly.decomposition.tucker(T, rank=(5,5,5), n_iter_max=10000)
O = einsum(C, U, V, W, 'a b c, i a, j b, k c -> i j k')
\end{lstlisting}

The restriction to isometric matrices is also usually relaxed in these decompositions (replacing the triangles with circles).

Sadly, these decompositions are not as well behaved as the SVD. Even determining the CP-rank of a tensor (the minimum number of nonzero singular values $\lambda_i $) is an NP-hard problem~\cite{hastad1990tensor} so the rank must usually be manually chosen rather than automatically found. Calculating these decompositions usually also requires iterative optimization, rather than just a simple LAPACK call.

\section{Tensor network decompositions}

Tensor networks are low-rank decompositions in exponentially-large dimensional spaces. The most common example is a Matrix Product State, also known as a tensor train.\footnote{These originally come from quantum physics. \textit{Matrix Product State} is the original name used by physicists,~\cite{fannes1992finitely, klumper1992groundstate, ostlund1995thermodynamic, vidal2003efficient, mcculloch2007from_dmrg_to_mps} while \textit{tensor train} is a more recent term sometimes used by mathematicians.~\cite{oseledets2011tensor}} It consists of a line of tensors, each tensor having one free ``physical'' leg, as well as ``bond'' legs connected its neighbors:
\[
    \includegraphics[height=20pt]{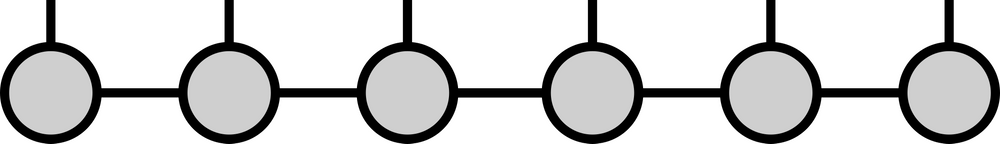} \;.
\]
Tensor trains are low rank decompositions of exponentially large dimensional spaces: contracting and flattening a tensor train produces an exponentially large dimensional vector:
\[
    \includegraphics[height=30pt]{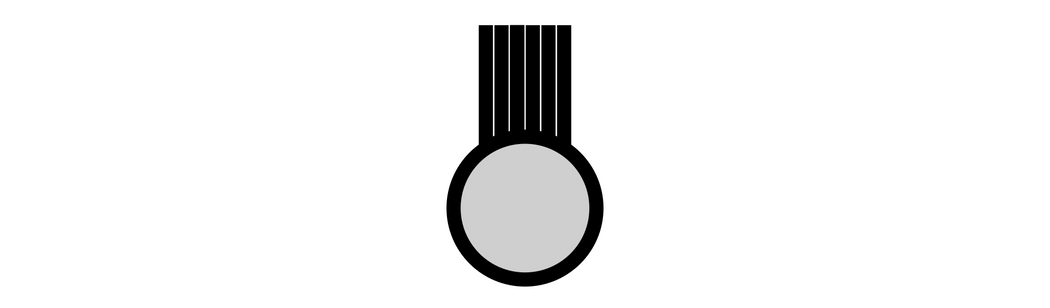} \;.
\]
A single large tensor like this can be decomposed into a tensor train using a series of SVDs:
\[
    \includegraphics[width=0.94\linewidth]{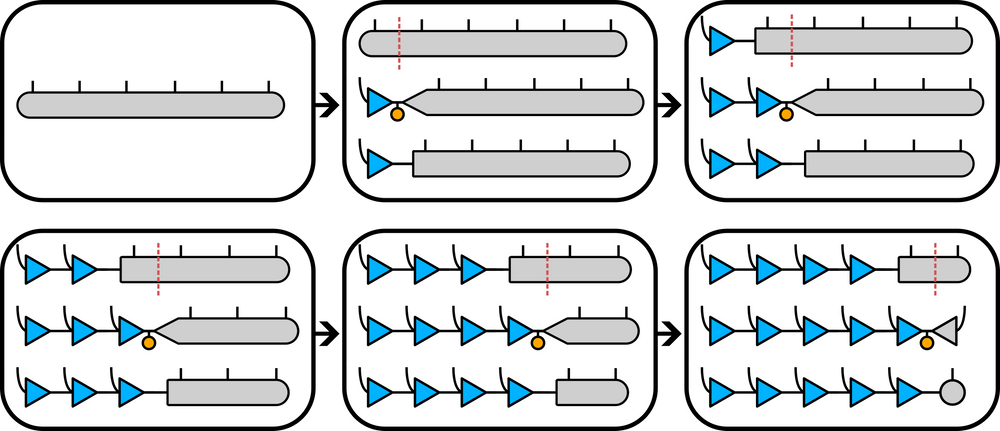} \;.
\]
 where each SVD needs only to be taken on the grey tensor rather than the whole network - the isometries in blue already form an orthonormal basis from the left, so taking an SVD of the grey tensor alone is equivalent to taking an SVD of the full network. In practice however, it's common to work directly in tensor network format from the start, rather than starting with a single extremely large tensor. Regardless, if all of the tensors are made isometric so that they point towards some spot,\footnote{For example, using a series of local SVDs and local contractions like so:\[\includegraphics[width=0.75\linewidth]{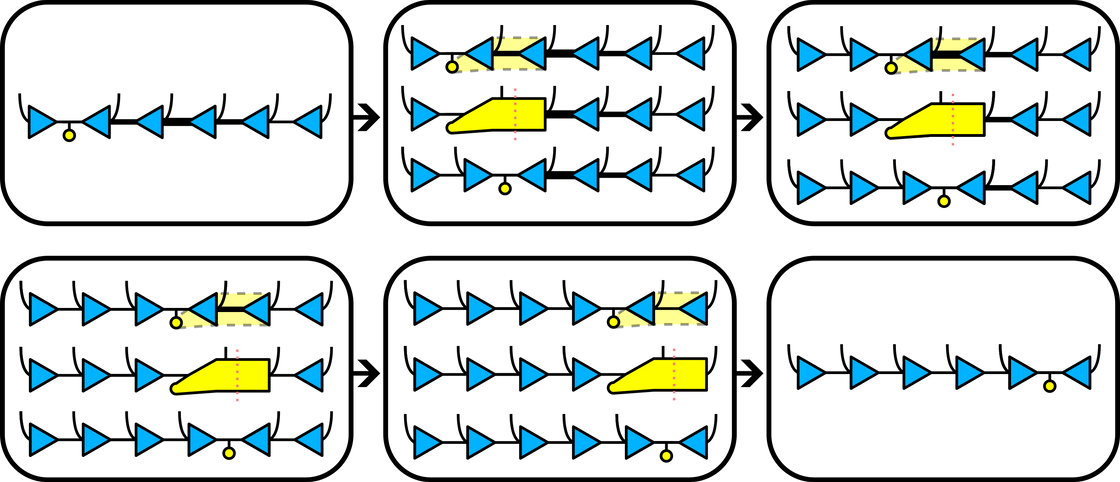}\]}  then the whole network is equivalent to an SVD around that spot:
\[
    \includegraphics[height=25pt]{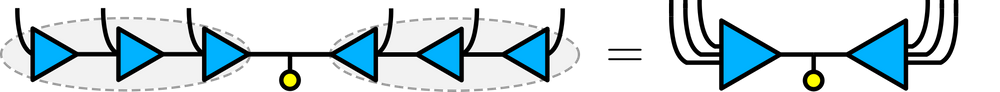} \;,
\]
and you get all of the benefits that come with the singular value decomposition, such as (sometimes) interpretable dominant singular vectors, optimal compression by discarding small singular values, and so on. Having a well defined orthogonality center like this also has many other advantages, such as fixing the ``gauge freedom'' in a tensor network.\footnote{Gauge freedom is the fact that many tensor networks contract to the same tensor, so transformations can be applied which affect the tensors, but don't affect what the network would contract to. For example, a resolution of the identity $I=XX^{-1} $ can be inserted on any bond. The matrix and its inverse can then be contracted into opposite surrounding tensors:This can even arbitrarily increase the bond dimension of the tensor network without changing what it represents, since the matrices $X $ and $X^{-1} $ can be rectangular.
    \[\includegraphics[width=0.25\linewidth]{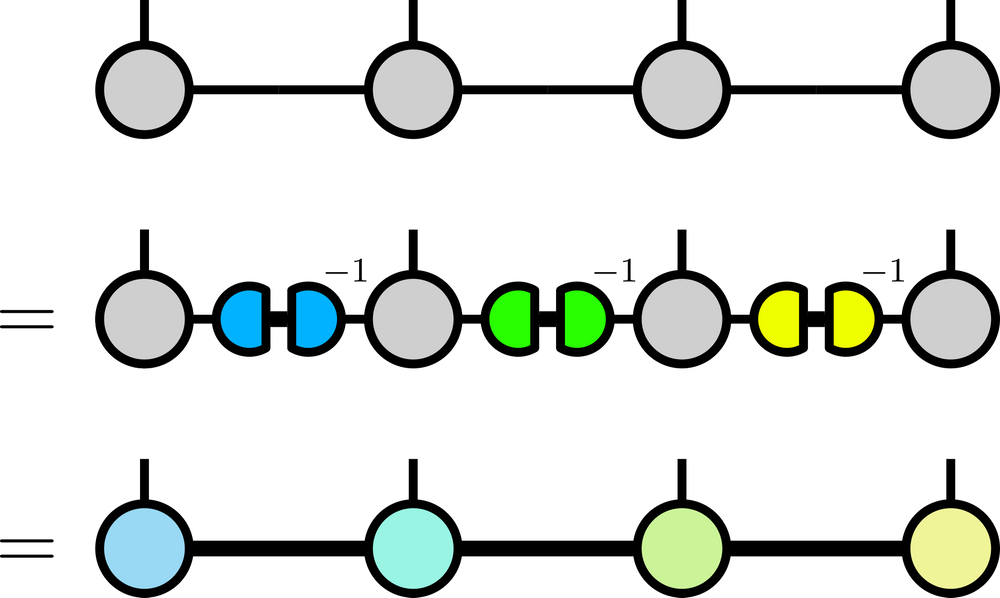} \;.\]
As a special case, the tensor network could even be multiplied by an entirely separate tensor network which contracts to the number 1:\\
    \[\includegraphics[width=0.5\linewidth]{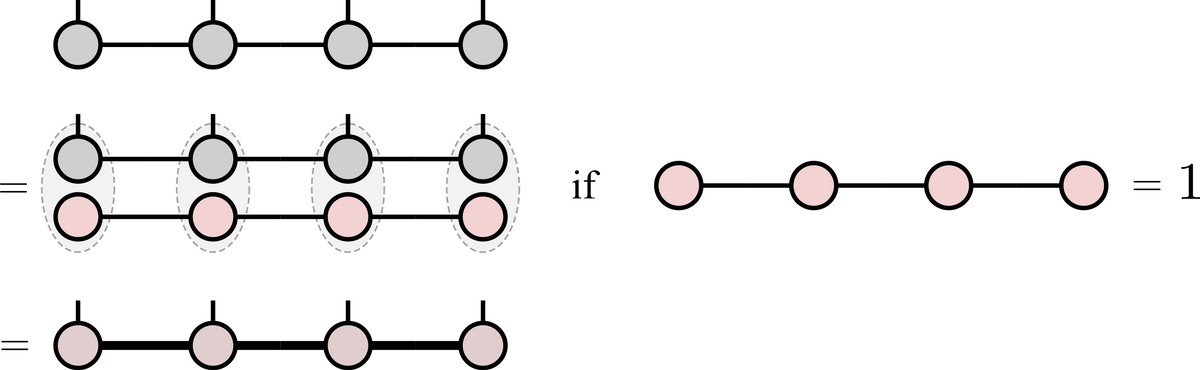} \;.\]
Making all tensors isometric towards some spot in the network will fix the gauge, since SVDs are unique (up to degeneracies in the singular value spectrum). Likewise, truncating the zero singular values will remove any unnecessarily large bond-dimensions. However SVDs only work for gauge-fixing in tensor networks without loops, such as tensor trains or tree tensor networks. }

Of course, this low-rank tensor train decomposition is only possible when only a few dominant singular vectors in each SVD are important - most singular values in each SVD must be small enough to be discarded. Otherwise, the bond dimension will grow exponentially away from the edges of the network. 

As a result, these decompositions work best when ``correlations'' between different sites far apart in the tensor network (that can't be explained by correlations with nearby sites) are relatively weak. In quantum physics for example, tensor networks are good at representing wavefunctions which don't have too much long-range entanglement.\footnote{Tensor networks work best at representing quantum states which have entanglement scaling with the surface area (rather than volume) of a subsystem~\cite{hastings2007area,masanes2009area,arad2013area,anshu2021area}}

There are many tensor networks commonly used in physics:
\begin{figure}[h]
    \centering
    \includegraphics[width=0.98\linewidth]{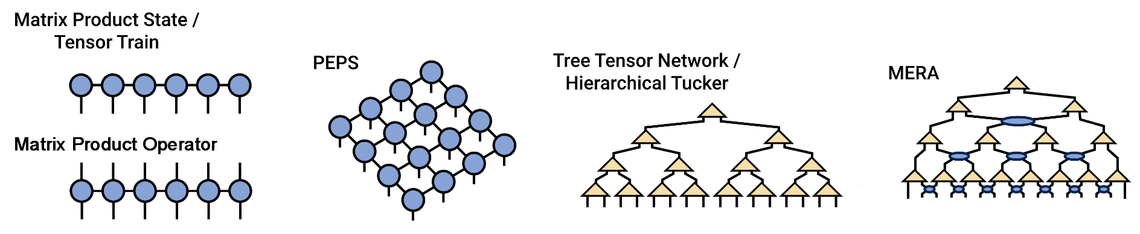}
    \caption{
Image adapted from \href{https://tensornetwork.org/}{tensornetwork.org}~\cite{stoudenmire2022tensornetworkorg} }
\end{figure}

\newpage
\section{Neural networks}\label{Neural networks}
The problem with tensor network notation is that it was developed for quantum physics, where \code{einsum} is all you need - no nonlinearities are allowed, \href{https://en.wikipedia.org/wiki/No-cloning_theorem}{not even copying}. So neural networks require going slightly beyond the standard graphical tensor notation, in order to represent nonlinearities.\footnote{Simon Verret already \href{https://simonverret.github.io/2019/02/16/tensor-network-diagrams-of-typical-neural-network.html}{has a post} on representing neural networks such as RNNs and LSTMs in graphical tensor notation, but I'll be using a different approach to the nonlinearities.}

\subsection{Dense neural networks }
Without nonlinearities, dense neural networks are equivalent to a bunch of matrices multiplied together - one for each weight layer. The data $x $ is input as a vector, which contracts with the matrices to yield the output vector $y $:\footnote{The bias terms can be accommodated into the matrices by concatenating \code{[1]} to the input vector $x $ and expanding the weight matrices appropriately.}
\[
    \includegraphics[width=0.38\linewidth]{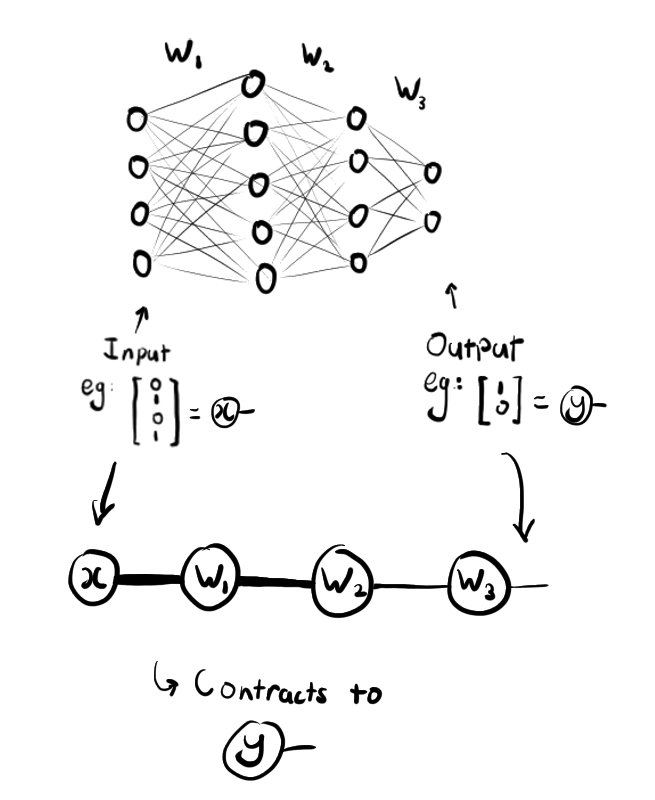} \;.
\]
Without nonlinearities, this contraction can be performed in any order. In fact it's equivalent to multiplying just a single weight matrix, as the contraction of weight matrices can be computed independent of the input vector $x $:
\[
    \includegraphics[width=0.34\linewidth]{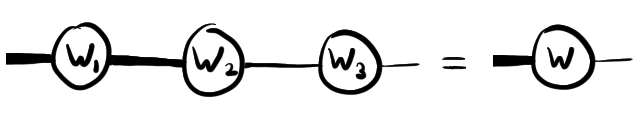} \;.
\]
Adding nonlinear functions, we introduce ``bubbles'': everything within the bubble must first be contracted, and then the nonlinear function can be applied to the single remaining tensor inside the bubble. As a result, these nonlinearities induce a fixed contraction order:
\[
    \includegraphics[width=0.4\linewidth]{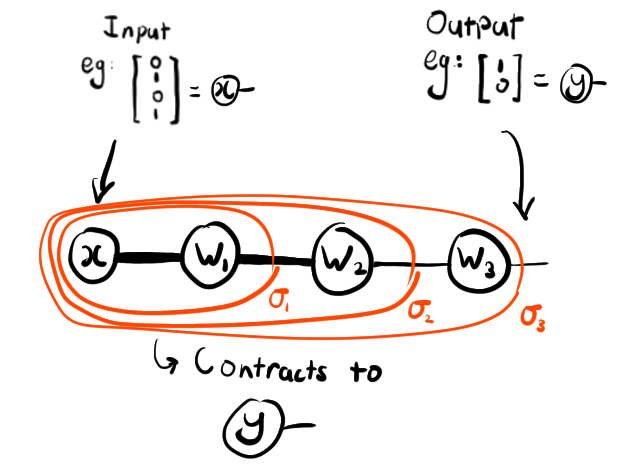} \;.
\]
First the input vector $x $ must be contracted with the first weight matrix $W_1 $, and only then can the elementwise nonlinearity $\sigma_1 $ be applied, and so on.

\newpage
\section{Transformers}

Transformers are being used for the largest and most capable AI systems, so they're an important focus for interpretability work. Transformers are traditionally used for sequence-to-sequence tasks, such as turning one string of words (or ``tokens'') into another. We'll explore them in graphical notation, with a focus on illustrating some of properties elucidated in \href{https://transformer-circuits.pub/2021/framework/index.html}{A Mathematical Framework for Transformer Circuits}.~\cite{elhage2021mathematical}

\begin{figure}
    \centering
    \includegraphics[height=0.91\textheight]{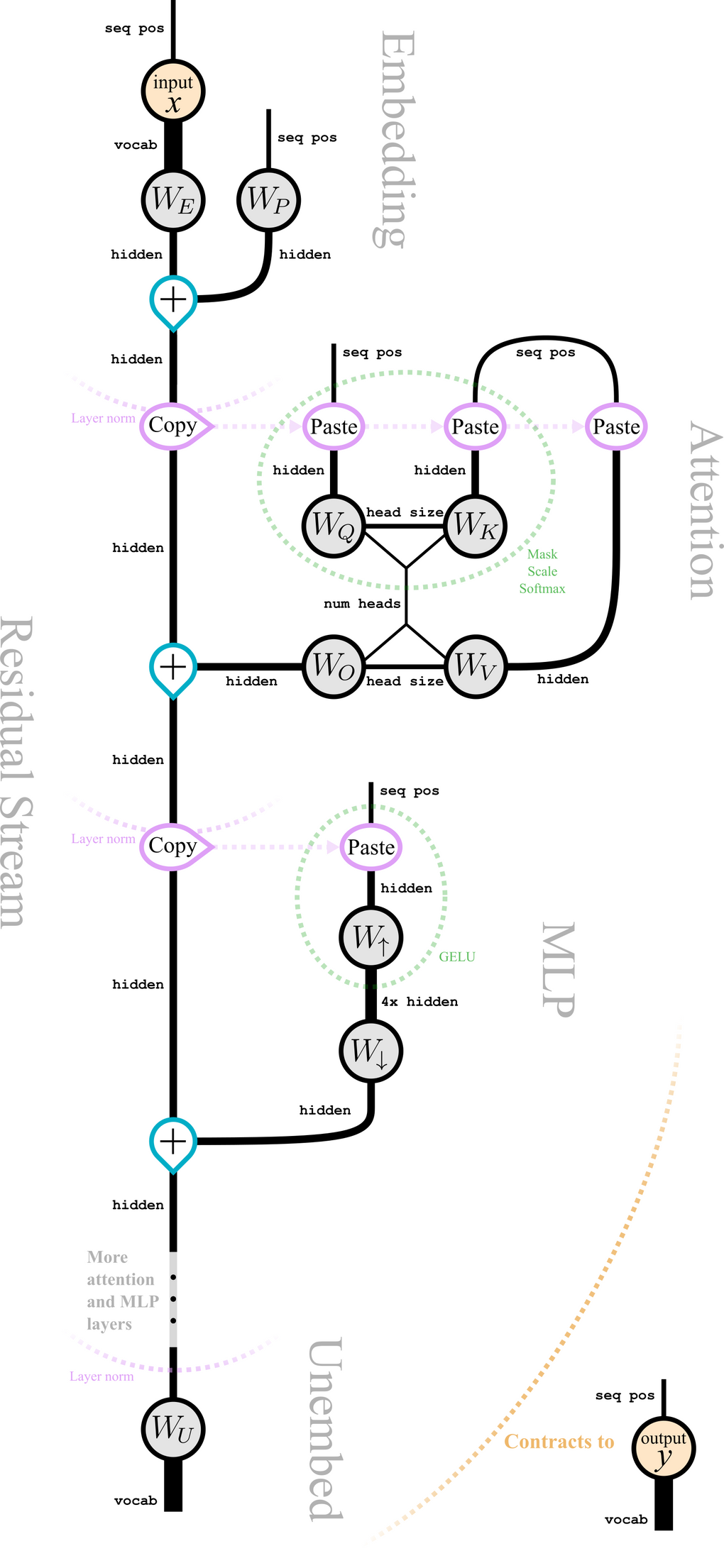}
    \caption{A tensor network diagram of GPT-2. Computationally, the network is contracted from top to bottom (input at the top, output at the bottom). Here we have shown one of the 12 layers of attention and MLP blocks, but the other layers are just repetitions of the attention and MLP blocks shown above, but with different learned weight parameters $W_Q$, $W_K$, $W_V$, $W_O$, $W_\uparrow$, $W_\downarrow$.}
    \label{fig:gpt2}
\end{figure}

So here's a tensor network diagram of GPT-2, with non-einsum operations shown in pink and green:
\[
\includegraphics[width=\linewidth]{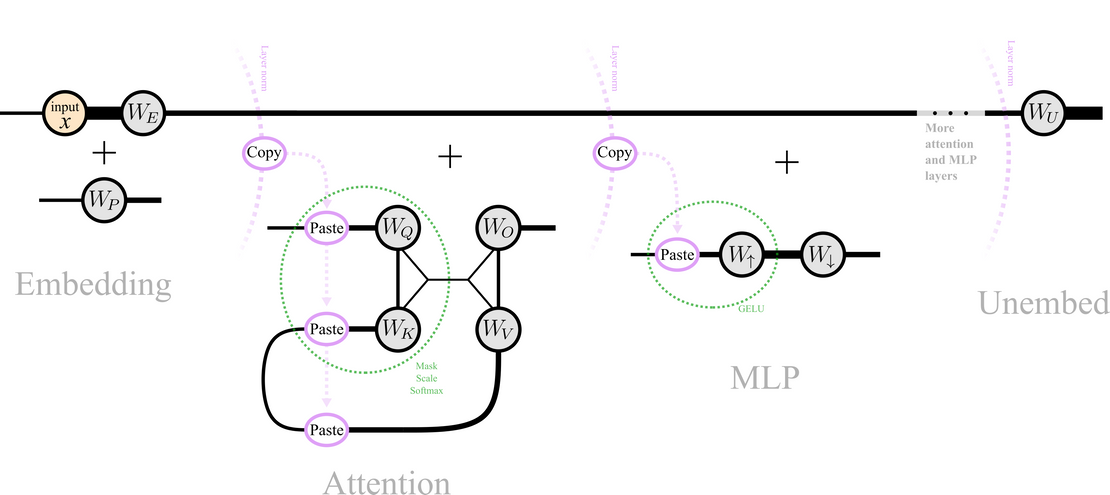}
\]
We see that the structure of the transformer is a series of distinct parts or ``blocks''. First, an embedding block, then ``Attention'' and ``MLP'' blocks alternate for many layers, then an unembedding block. Figure \ref{fig:gpt2} shows this diagram flipped around vertically so we have more room to label the legs and see what's going on. 
Here we've also changed how we denote the elementwise addition of tensors (in blue), to emphasize the most dominant part of a transformer: the vertical ``backbone'' on the left known as the residual stream. This is the main communication channel of the network. Each layer copies out information from the residual stream, uses it, and then adds new transformed information into (or subtracts it out of) the residual stream. 

Here are the dimensions and descriptions of each leg in the above diagram for GPT-2: 
\begin{table}[h]
\centering
\begin{tabular}{p{0.12\linewidth} p{0.20\linewidth} c p{0.55\linewidth}}
Leg & Dimension in GPT-2 & ~ &Description \\
\hline
\code{{ seq pos }} & up to context length (1024) & ~ &The number of tokens in the input text. (Indexes which token in the input text) \\
\code{{ vocab }} & 50257 & ~ &The number of tokens in the vocabulary of all possible tokens (Indexes tokens by their token ID). \\
\code{{ hidden }} & 768 & ~ &The dimension of the residual stream on each token (space for information stored on that token) \\
\code{{ num heads }} & 12 & ~ &The number of attention heads per attention layer \\
\code{{ head size }} & 64 & ~ &The compressed dimension of each attention head  \\
\end{tabular}
\end{table}

\subsection{Input and output}

The input data \includegraphics[height=20pt]{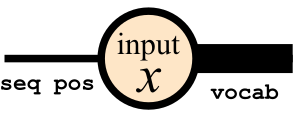} can be seen as a map from each position in the input text to the ID of the token at that position: $x =\delta_{\text{seq\_pos},\; \text{corresponding\_token\_ID}} $. Like most delta-tensors it's computationally cheaper to use clever indexing rather than actually representing it as a tensor, but here's what it looks like regardless:
\[
\includegraphics[width=0.85\linewidth]{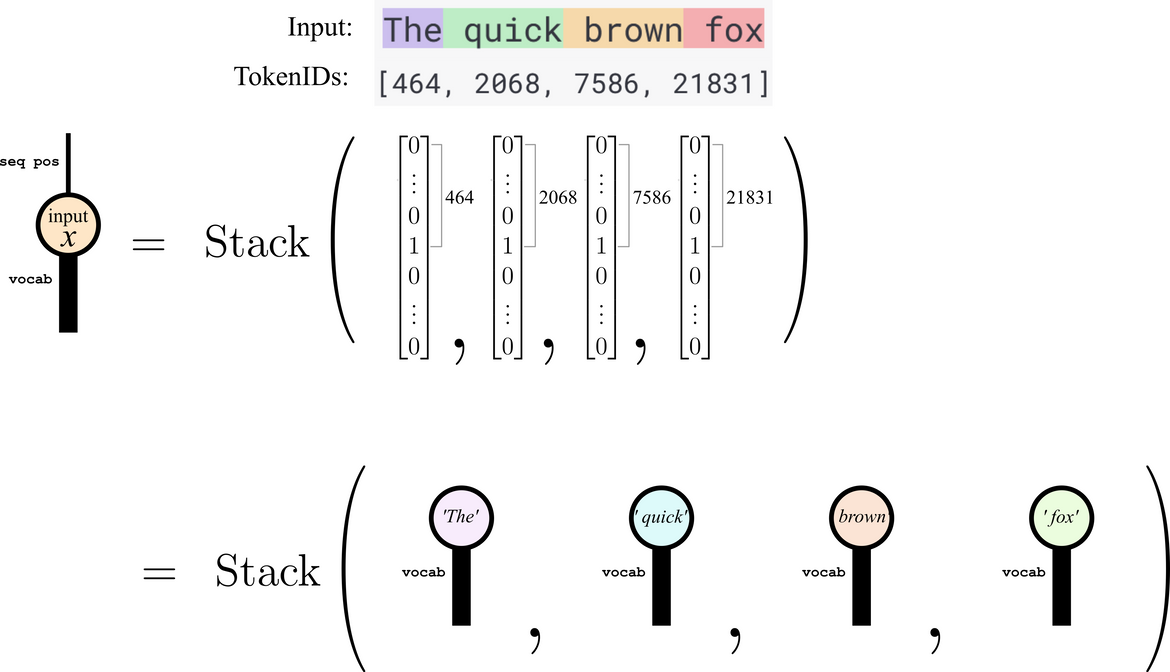}\;.
\]
The output tensor $y $ (found by contracting the input with the whole network) has the same shape, but in general every entry will be nonzero, as it represents the log probability that the model predicts for every possible next token at every position. The vector at the final token position represents the log probability distribution for the unknown next word, which can be sampled to turn the predictive model into a generative model.

The embedding layer is responsible for getting the input into the model, by compressing (projecting) the input vector on each token from a $\dim($\code{{vocab}}$\approx50,000 $ dimensional space down to a $\dim( $\code{{hidden}}$)\approx768 $ dimensional space known as the residual stream of each token:
\[
\includegraphics[width=0.85\linewidth]{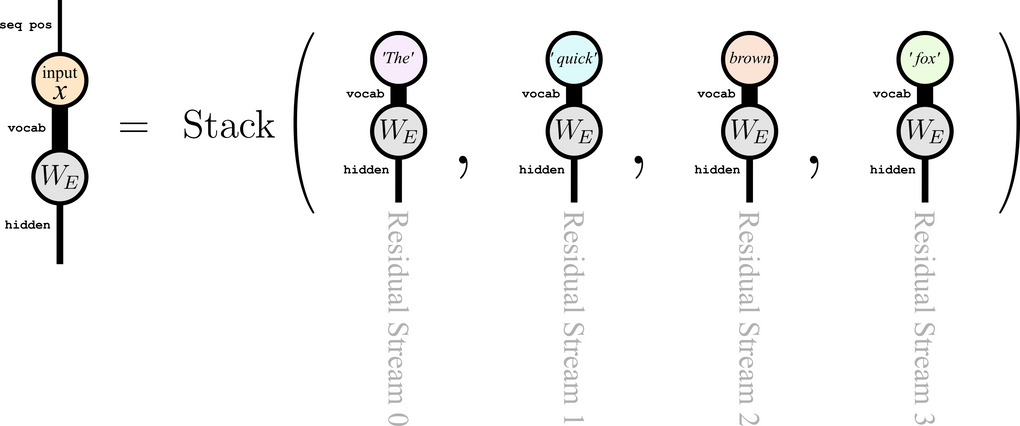}\;.
\]
The total dimension of the residual stream is $\dim( $\code{{seq pos}}$)\times \dim( $\code{{hidden}}$) $: there is one stream for each input token, and each token's stream is of dimension $\dim( $\code{{hidden}}$) $. 

Presumably this process of embedding involves creating a vector representing the token's meaning independent of any context from the tokens around it, by packing similar tokens into similar parts of residual-stream space (though \href{https://dynalist.io/d/n2ZWtnoYHrU1s4vnFSAQ519J\#z=3br1psLRIjQCOv2T4RN3V6F2}{superposition} may also be involved). 

The embedding also has another component: the positional embedding. This is a simple fixed vector for each position, added to each residual stream independent of the token at that position. It lets each residual stream have some information about where in the sequence it is located:
\[
\includegraphics[width=0.85\linewidth]{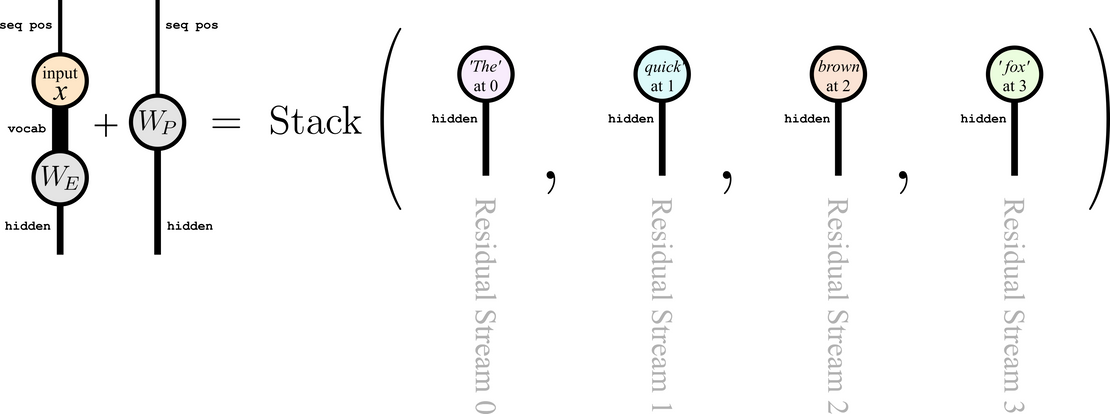}\;,
\]
as long as a subspace in the residual stream is reserved to store and use this positional information.

However different transformer architectures tend to use very different kinds of positional embedding techniques, so for simplicity throughout the rest of the sections we'll contract the input and the embeddings into a single tensor:
\[
\includegraphics[width=0.55\linewidth]{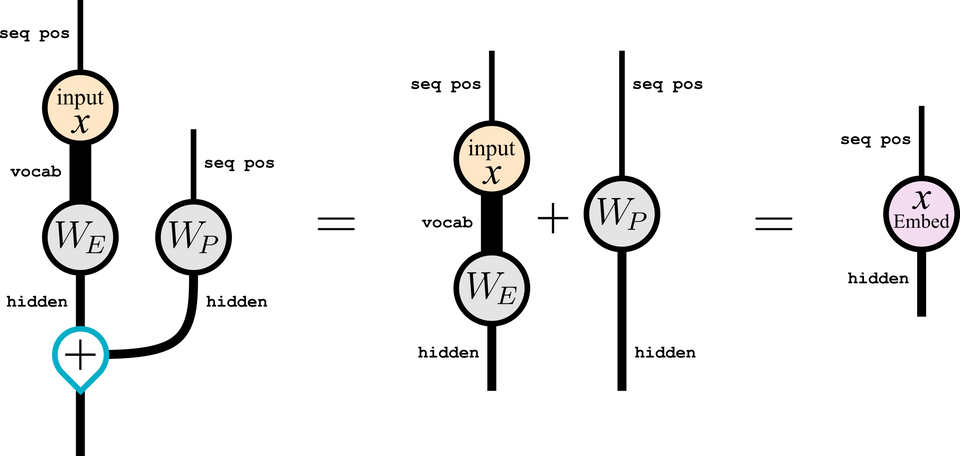}\;,
\]
but feel free to substitute in the expanded form whenever you like.

The unembedding matrix works similarly, but without any positional component.

The structure of the transformer from here is a series of ``blocks'' which copy from, then add back into the residual stream. There are two kinds of blocks: Attention and MLP, which alternate down through the layers. 

\subsection{MLP}

The MLP layer is most like a dense neural network, though it is residual (copying from and adding into the residual stream, rather than modifying it directly), and acts only on the \code{{hidden}} index. Unlike an MLP layer, a general transformation would be able to move information between token positions, by also acting on the \code{{seq pos}} index like so:
\[
\includegraphics[width=0.18\linewidth]{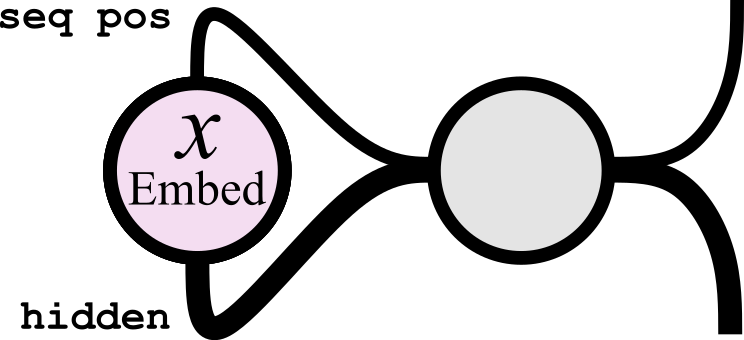}\;,
\]
but MLP layers in transformers only involve contractions onto the \code{{hidden}} leg, and so act independently on each token, meaning that they can't move information from one residual stream to another: 
\[
\includegraphics[width=0.95\linewidth]{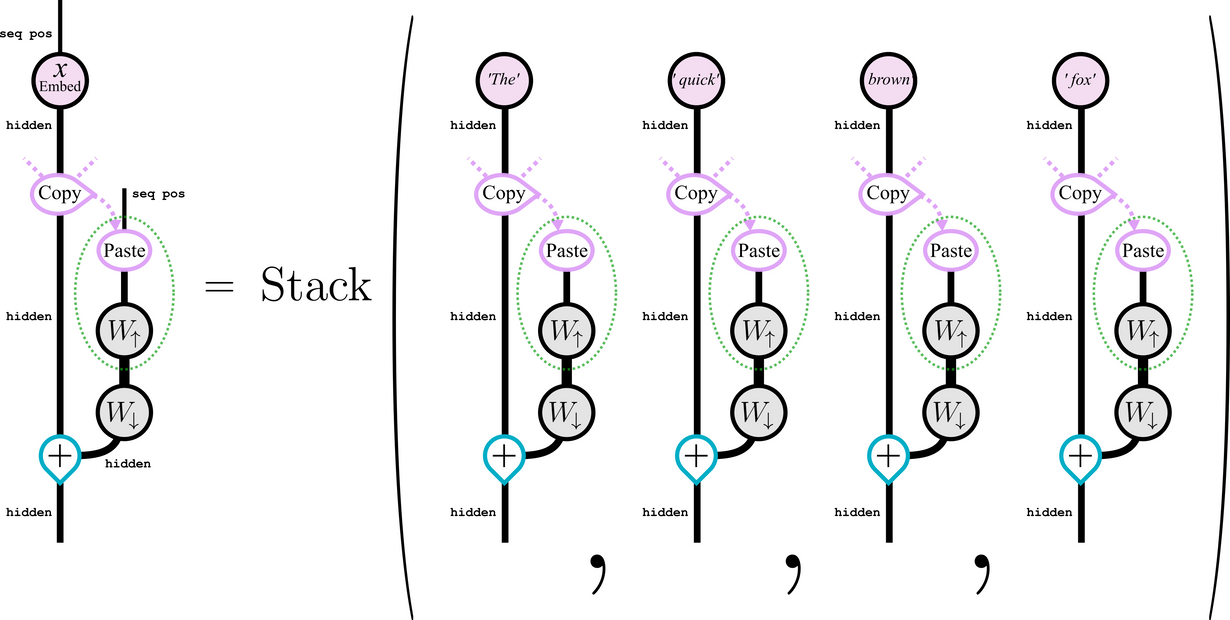}\;.
\]

Despite acting independently on each residual stream, MLPs make up the vast majority of the parameters in a transformer because they project up to a higher-dimensional space with the $W_\uparrow $ matrix\footnote{a $4\times $ higher dimensional space in the case of GPT-2} before applying a GELU nonlinearity and projecting back down again with $W_\downarrow $. The MLP parameters in $W_\uparrow $ and especially $W_\downarrow $ seem to be where most of the trained facts and information are stored, as evidenced by \href{https://aclanthology.org/2021.emnlp-main.446/}{Transformer Feed-Forward Layers Are Key-Value Memories}~\cite{geva-etal-2021-transformer}, the \href{https://rome.baulab.info/}{ROME (Rank One Model Editing) paper}~\cite{meng2022locating} and subsequent work on \href{https://dynalist.io/d/n2ZWtnoYHrU1s4vnFSAQ519J\#z=qeWBvs-R-taFfcCq-S_hgMqx}{activation patching} for locating and editing facts inside transformers.

\subsection{Attention}

Each attention layer consists of a number of heads, which are responsible for moving relevant data from the residual stream of preceding tokens, transforming it, and copying it into the residual stream of later tokens. Each head acts effectively independently - we can see this in our diagram by replacing the \code{{num heads}} leg with a sum, and seeing that each term contributes independently to the residual stream:
\[
\includegraphics[width=0.80\linewidth]{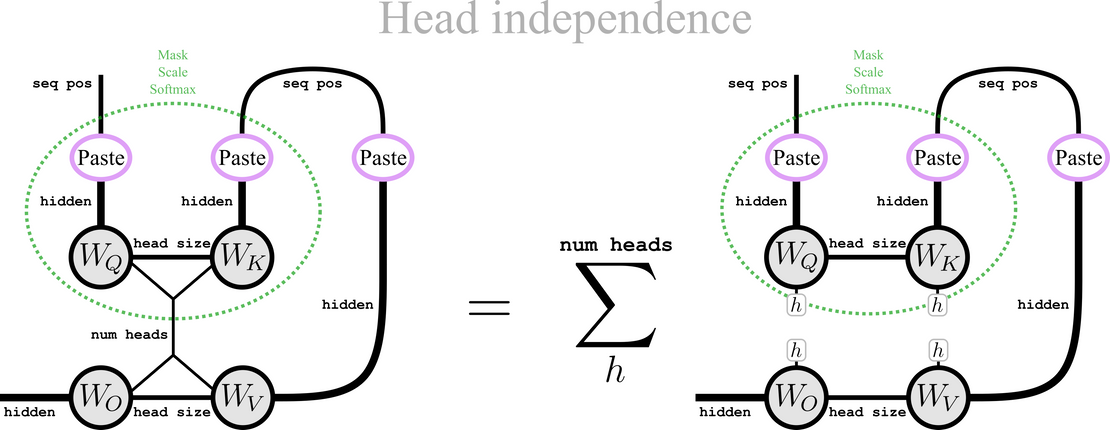}\;.
\]

Attention heads host the most easily interpretable parts of a transformer: the attention patterns. These are low-rank matrices (one matrix for each attention head, as indexed by the \code{{num heads}} leg) calculated like so:
\[
\includegraphics[width=0.55\linewidth]{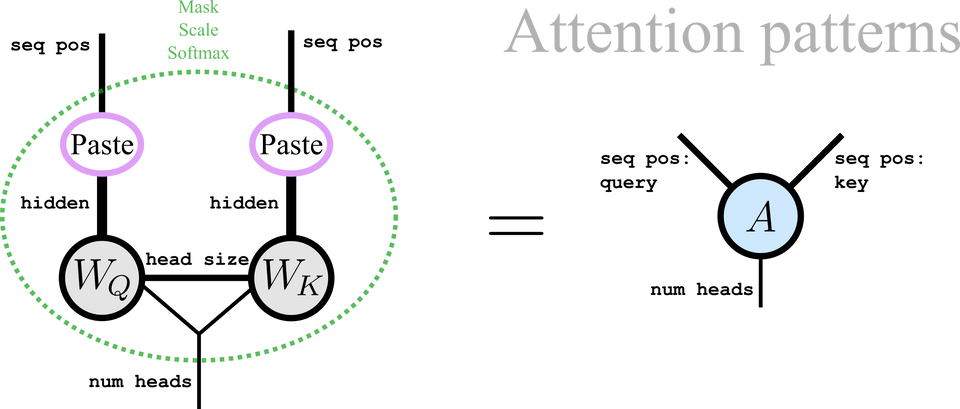}\;.
\]

Attention patterns determine how information is moved between tokens. For example, this attention pattern seems to move information from earlier tokens matching a pattern onto tokens immediately preceding the equivalent token in a new language:\footnote{This specific ``induction head'' kind of attention pattern will only be seen after the first layer, because it must arise as a result of composition with attention head(s) in previous layers.}
\[
\includegraphics[width=0.47\linewidth]{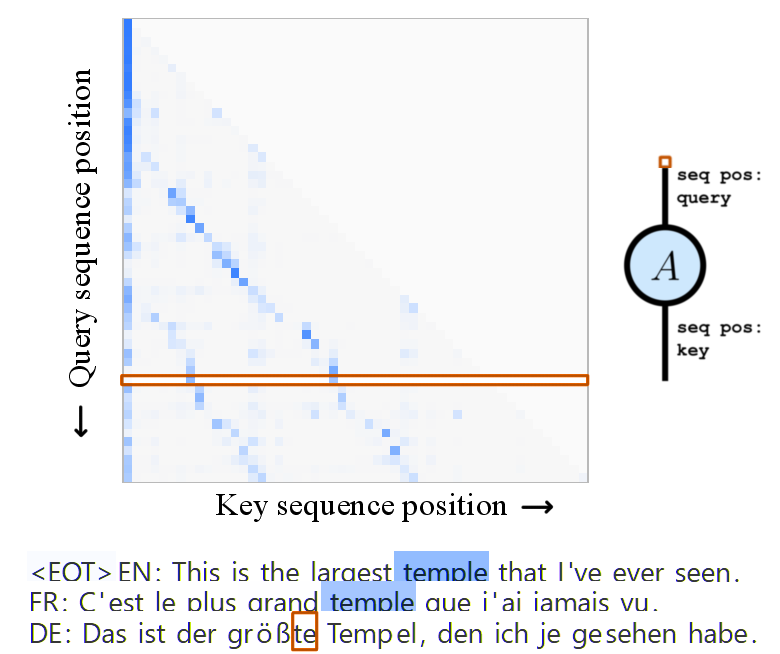}\;.
\]

You can play with \href{https://transformer-circuits.pub/2022/in-context-learning-and-induction-heads/index.html\#performing-translation}{the interactive version} of this in the \href{https://transformer-circuits.pub/2022/in-context-learning-and-induction-heads/index.html\#performing-translation}{In-context Learning and Induction Heads paper}.~\cite{olsson2022context} We'll see an example of how an attention pattern like this can come about in section \ref{sec:toy_induction}, but for now we can just take attention patterns for granted. 

Rather than computing attention patterns on the fly based on the current context in the residual stream (the two ``pasted'' tensors in pink), the attention pattern can be ``frozen'' for easier interpretability, fixing the $A $ tensor and therefore fixing a specific pattern of information movement. When this is done, the attention block simplifies to
\[
\includegraphics[width=0.73\linewidth]{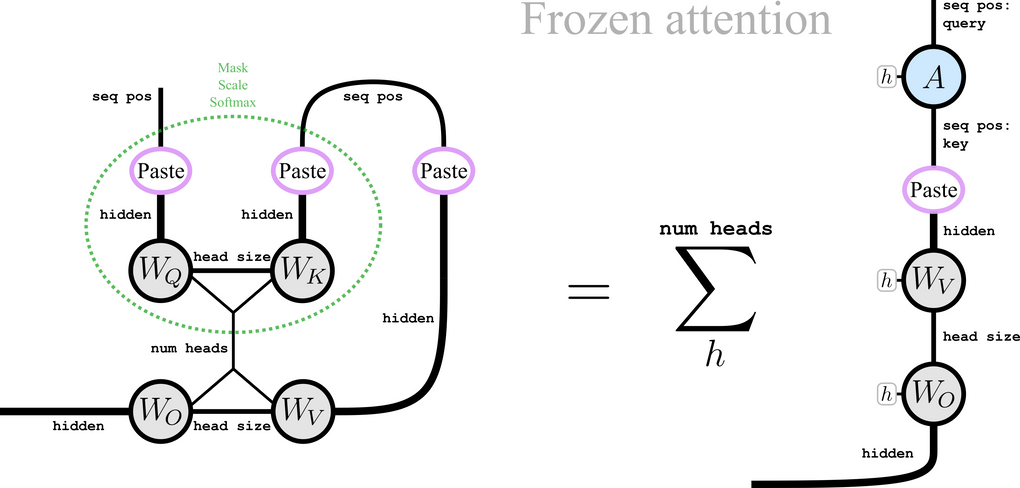}\;,
\]
which is completely linear: the only non-einsum operation is now just a single copy and paste from the residual stream.\footnote{Copying and pasting can in general be nonlinear, for example if products are taken between copies of an object, that's the same as raising it to some power: a nonlinear operation. But here with attention frozen there are no products taken between copies: just a sum when the attention result is added back into the residual stream.} We can also see that the attention pattern $A $ is the only transformation in the whole network which ever acts on the \code{{seq pos}} index: every other tensor is contracted into the \code{{hidden}} index, so every other linear transformation can be described independently for each token. This is why the attention pattern is the only part of the network that can move data between tokens. 

With attention frozen, we can represent the sum over heads in a number of ways:
\[
\includegraphics[width=0.95\linewidth]{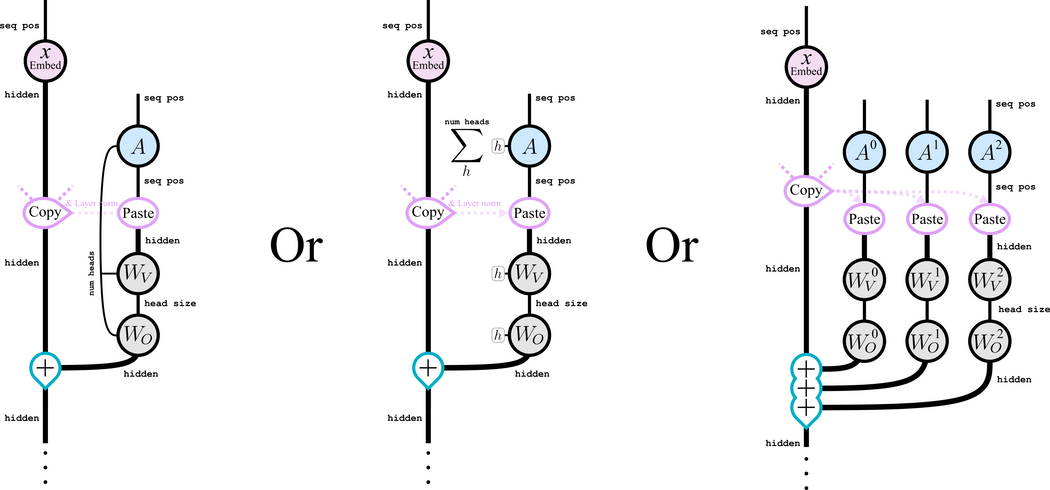}\;,
\]
these are all equivalent, just placing different emphasis on the independence of each head. You can even go in the opposite direction and emphasize the matrix-multiply nature of the \code{{num heads}} leg, reminiscent of low rank decompositions like SVD but for operators:\footnote{This is reminiscent of other decompositions of tensor operators into sums of rank-one tensor products, such as sums of strings of single-site Pauli operators in quantum error correcting codes, and Matrix Product Operators (MPOs) more generally. }
\[
\includegraphics[width=0.42\linewidth]{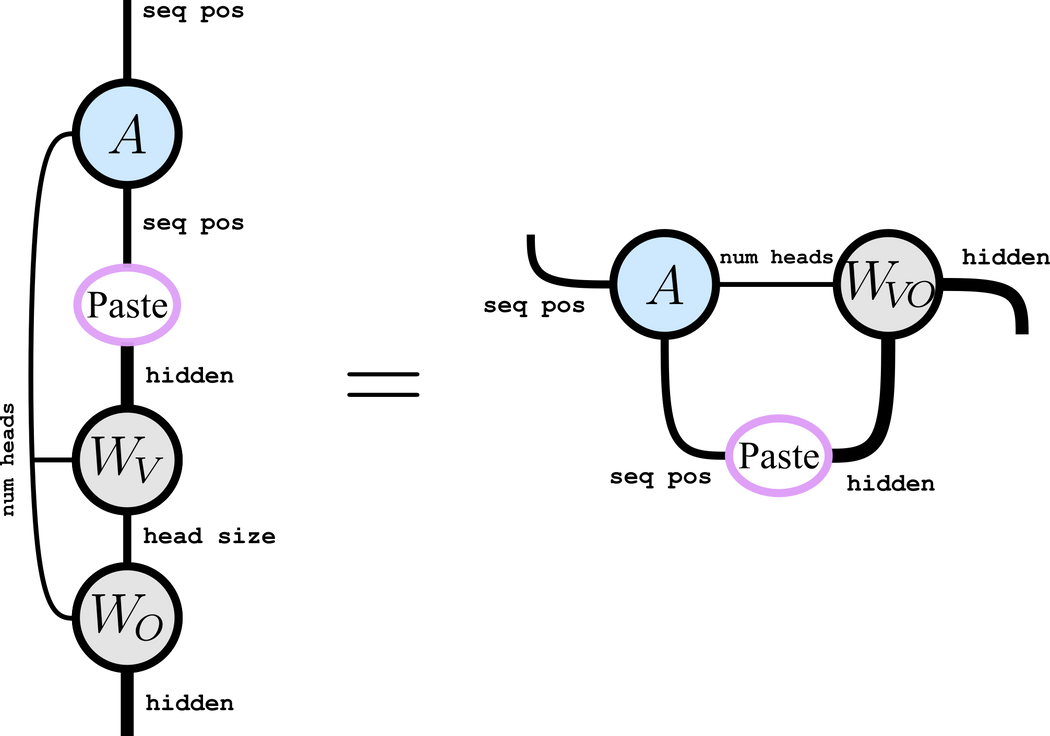}\;,
\]
though this low-rank-operator way of looking at attention is probably less useful than the sum-of-heads way of looking at it. 

\newpage
\section{Composition and path expansion}
Layers of Attention and MLP blocks don't just act in isolation: they all copy from and write to the same residual stream, so later layers can use information computed in earlier ones. Still following \href{https://transformer-circuits.pub/2021/framework/index.html}{A Mathematical Framework for Transformer Circuits},~\cite{elhage2021mathematical} I'll ignore MLP blocks and focus only on the ways that attention heads in an earlier layer can ``compose'' with those in a later layer. I'll also ignore small but annoying nonlinearities like \href{https://transformer-circuits.pub/2021/framework/index.html\#technical-details}{layer normalization}.

So here's a two layer attention-only transformer:
\[
\includegraphics[width=0.99\linewidth]{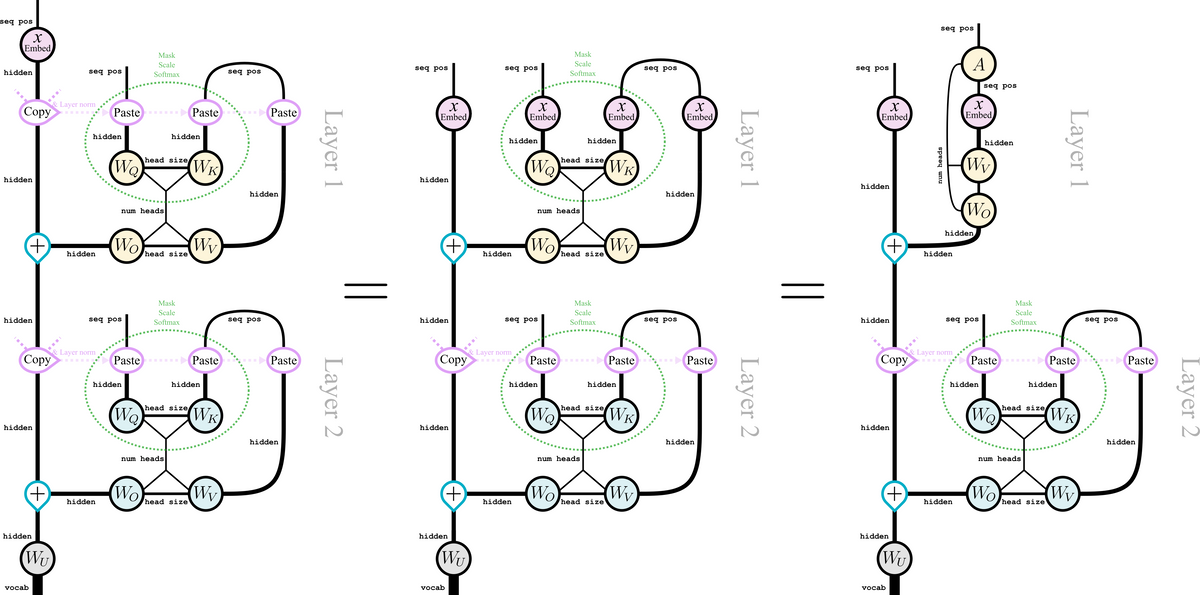}\;,
\]
where in the middle we've gone ahead and copied the input into the first attention layer, and on the right we've contracted the attention patterns in the first layer.

When the second layer copies from the residual stream, it will copy the a sum of terms from earlier layers. Rather than treating the result of this sum as a single complicated object, we can keep the sum expanded as two separate terms: the original input $x $ plus a ``perturbation'' caused by the first layer:
\[
\includegraphics[width=0.55\linewidth]{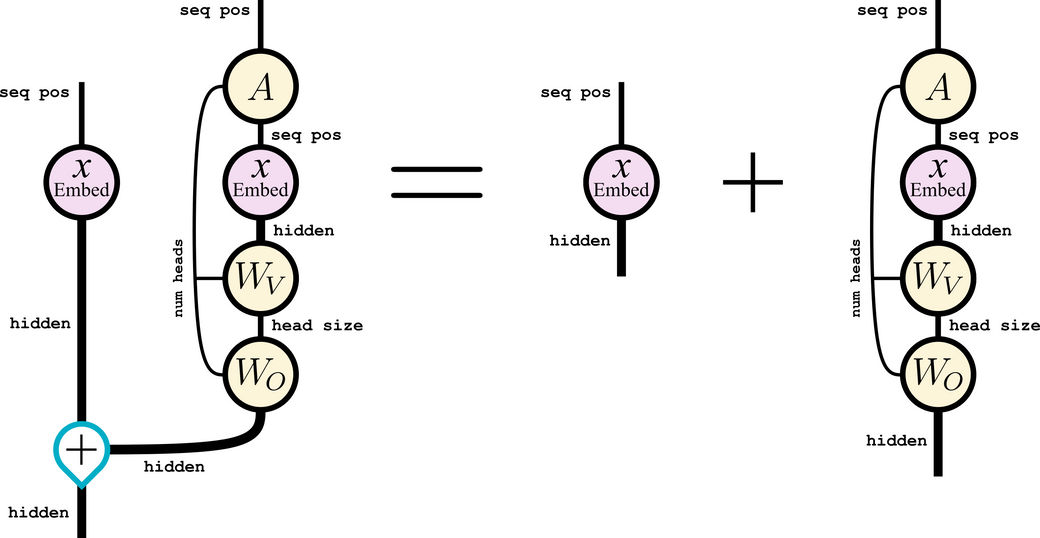}\;,
\]

Now we can expand the output of the whole network as a sum of terms like this, in something reminiscent of a \href{https://en.wikipedia.org/wiki/Perturbation_theory}{perturbation theory}:
\[
\includegraphics[width=0.95\linewidth]{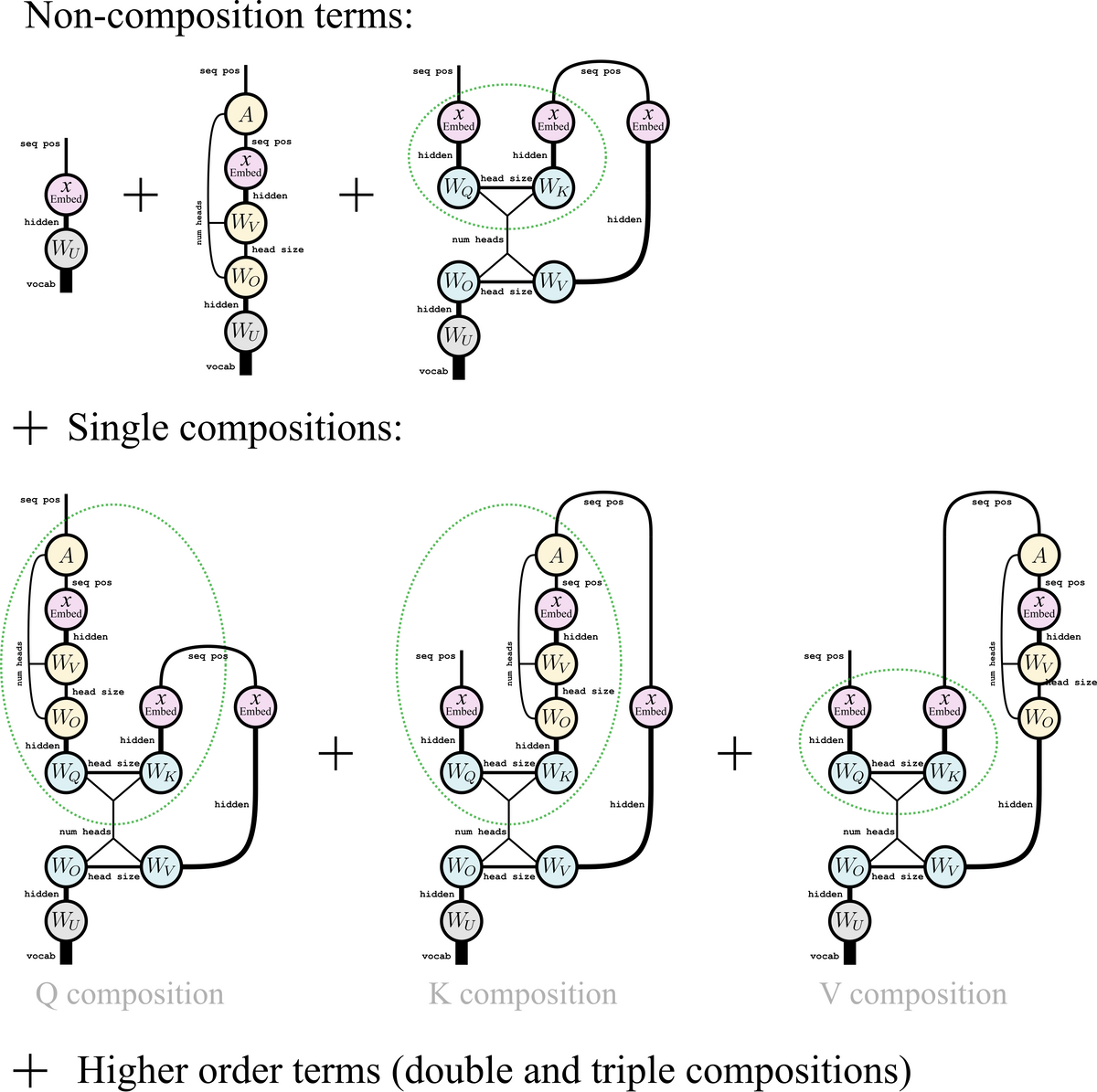}\;,
\]
Note: this expansion is not technically true as written due to the nonlinearities (green ellipses) not acting independently on each term in the sum: this is what it means to be nonlinear. However these nonlinearities are pretty linear - consisting only of a mask (which is linear), a re-scale, and a softmax to ensure that the elements in each row of the attention pattern add to 1. The layer normalization nonlinearities (not shown) \href{https://transformer-circuits.pub/2021/framework/index.html\#technical-details}{can also largely be treated as approximately linear} with sufficient care.~\cite{elhage2021mathematical} So we see that there are three simplest kinds of nontrivial attention composition: Q-composition, K-composition, and V-composition. We can simplify the V-composition and non-composition terms slightly by noticing when the attention patterns in the second layer can also be frozen:
\[
\includegraphics[width=0.90\linewidth]{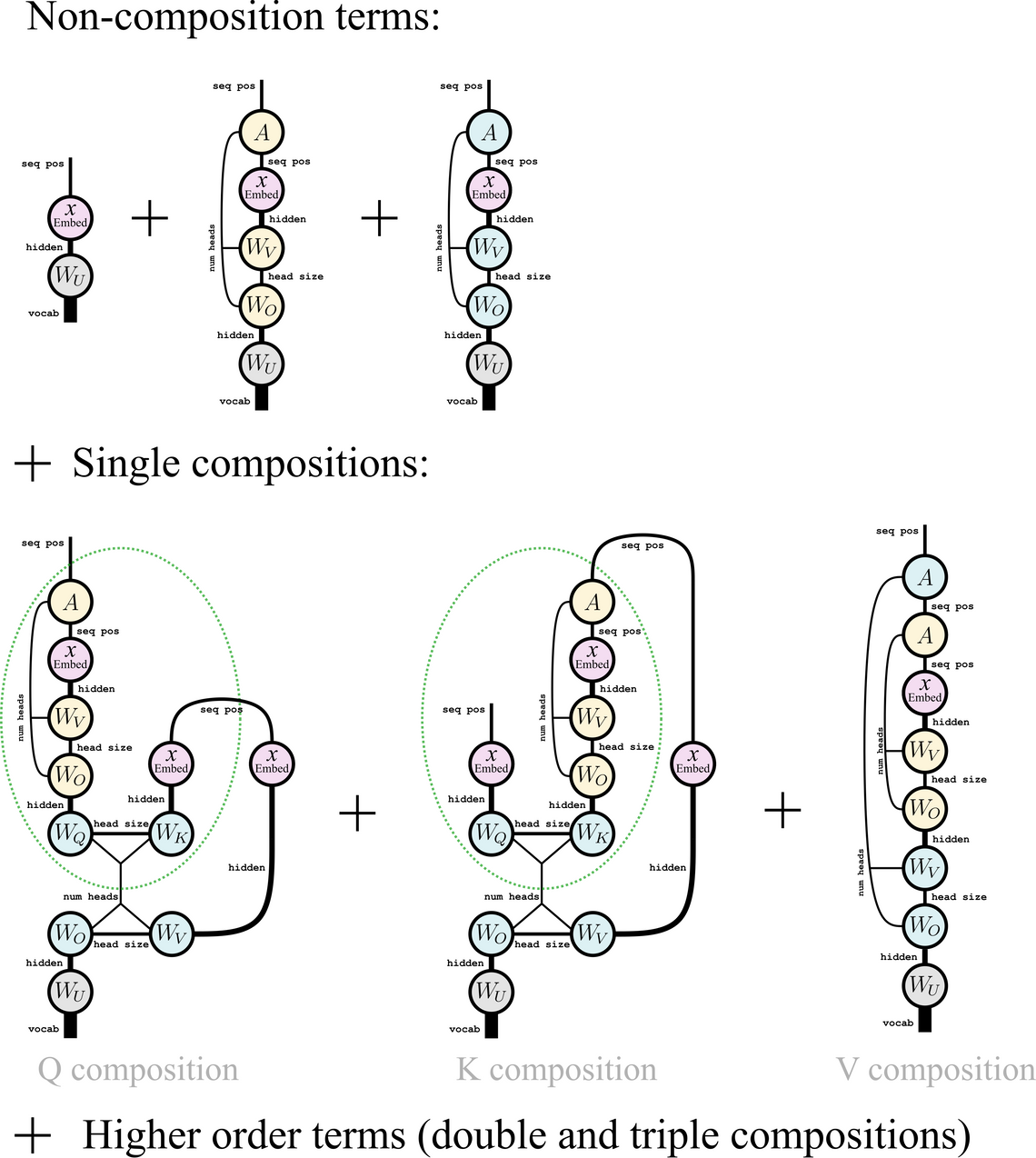}\;,
\]
so we can see that V-composition has a simpler iteration structure than Q or K composition - the attention pattern is just formed by matrix-multiplying two attention patterns, and likewise for the $W_{O} $ and $W_V $ transformations.

There are ten terms which contribute to the final answer for two layers: three non-compositions (shown), three single compositions (shown), and four higher-order compositions (not shown). But we can (and usually should) also expand out the \code{{num heads}} indices as sums, and have a separate term for each head and combination of heads. Likewise for the \code{{seq pos}} indices if we want to consider the contributions to or from specific token positions. \href{https://transformer-circuits.pub/2021/framework/index.html\#additional-intuition}{MLP layers could also be incorporated into this expansion sum},~\cite{elhage2021mathematical} though their stronger nonlinearities would require some kind of linearization, and they only have one kind of nontrivial composition anyway. Regardless, the number of terms grows exponentially with the number of layers, so this kind of trick will only be useful if we have some reason to think that most of the work is being done by relatively few terms (preferably low order ones). 

Some intuition for thinking that relatively few terms are important comes from noticing that each head can only write to a relatively small subspace of the residual stream, because the \code{{head size}} dimension is small compared to the \code{{hidden}} dimension, and so each head is putting a relatively small dimensional vector into a relatively high dimensional space with $W_O $ and selecting small parts of a high dimensional residual stream with with $W_Q $ and $W_K $, leaving sufficient room in the residual stream for most heads in different layers to act mostly independently if they want.\footnote{Though the same isn't true for MLP layers. Additionally, if \code{{num heads}} $\times $\code{{head size}} $\approx $ \code{{hidden dim}} (as is usually true), and the contribution of each head to its subspace is not small, then some decent number of heads per layer \textbf{must} interact with heads in the previous layer.} Of course its an empirical matter if low order terms \textit{actually }explain most of the relevant behaviors, so this should be \href{https://transformer-circuits.pub/2021/framework/index.html\#term-importance-analysis}{checked empirically}. There may also be much more effective ways of decomposing the computations of a transformer into a series of terms or circuits like this, such that more of the relevant behavior is explained by fewer more interpretable terms. One recent method attempting to find them is the \href{https://arxiv.org/abs/2304.14997}{Automated Circuit DisCovery (ACDC) }algorithm.~\cite{conmy2023towards} 

\section{Example: toy induction head}\label{sec:toy_induction}

We'll finish off by constructing one toy example of an induction head: a circuit detecting what should come next in a repeated string of tokens. For example, consider predicting the next word in some text like ``The quick brown fox [...] the quick brown''. It seems like ``~fox'' should come next to fit the pattern. Information from the earlier ``~fox'' token should be copied into the final ``~brown'' token's residual stream, so that the model can predict ``~fox'' for the next word there:
\[
\includegraphics[width=0.98\linewidth]{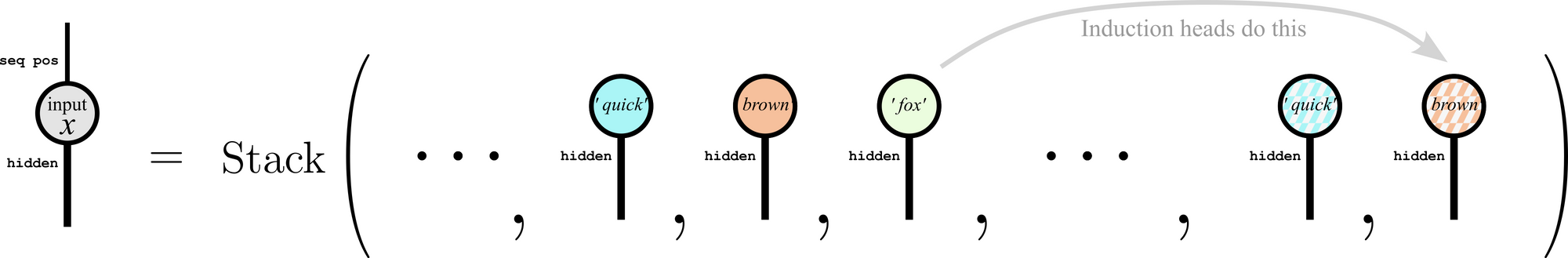}\;,
\]
This is known as induction, which is a type of in-context learning. There are many ways to make these induction heads, and real induction heads are likely to be messy, but things resembling them \href{https://transformer-circuits.pub/2022/in-context-learning-and-induction-heads/index.html}{have been found} in models of virtually all sizes.~\cite{olsson2022context} We'll construct a handcrafted toy example of an induction circuit, by forming a ``virtual induction head'' from the K-composition of two heads in different layers:
\[
\includegraphics[width=0.28\linewidth]{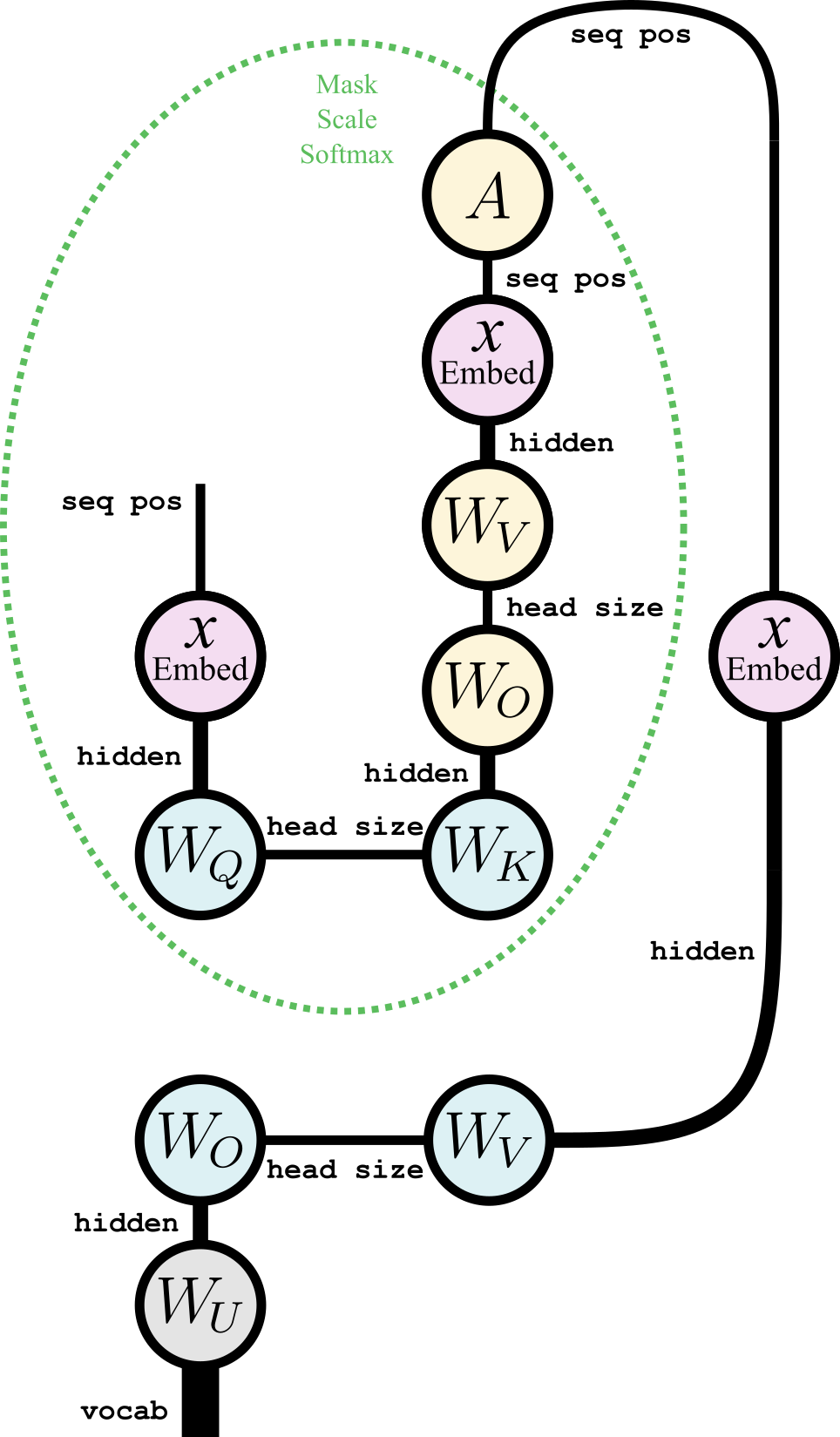}\;,
\]
Everything in this diagram is now a matrix, so we could draw it to emphasize that it's just a sequence of matrix multiplications:
\[
\includegraphics[width=0.7\linewidth]{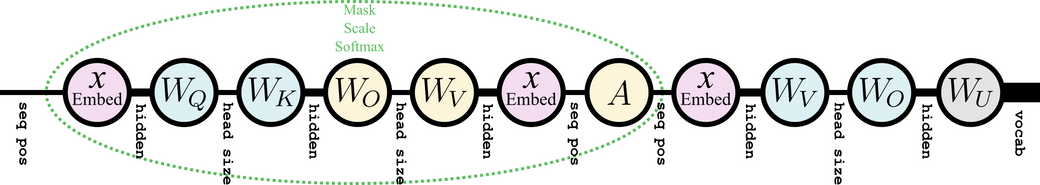}\;,
\]
but we'll stick with the previous format so you can more easily see how the circuit fits into the rest of the network. 

In order for this to act like an induction head, the last ``~brown'' input token (\includegraphics[height=20pt]{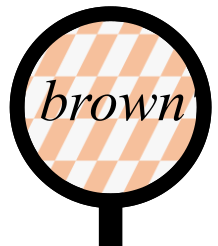}) should output a ``~fox'' token:\footnote{The original ``~brown'' vector will probably also have to be subtracted out of the residual stream somewhere too. This is also ignoring positional embeddings.} 
\[
\includegraphics[width=0.47\linewidth]{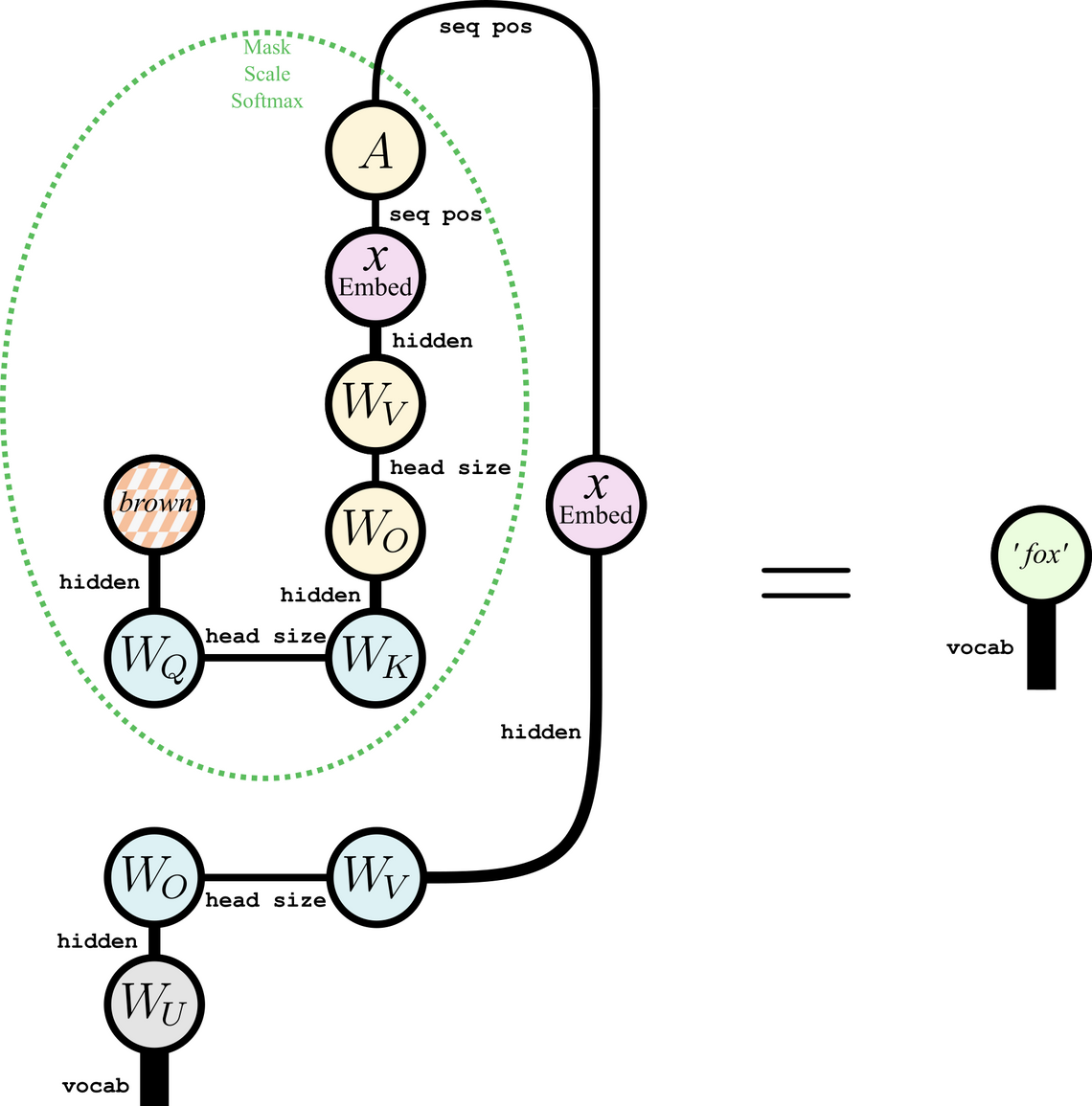}\;.
\]
Induction involves exploiting repeated patterns, so induction heads had better be able to pattern match. Here's the subcomponent of our induction circuit which will do that:
\[
\includegraphics[width=0.77\linewidth]{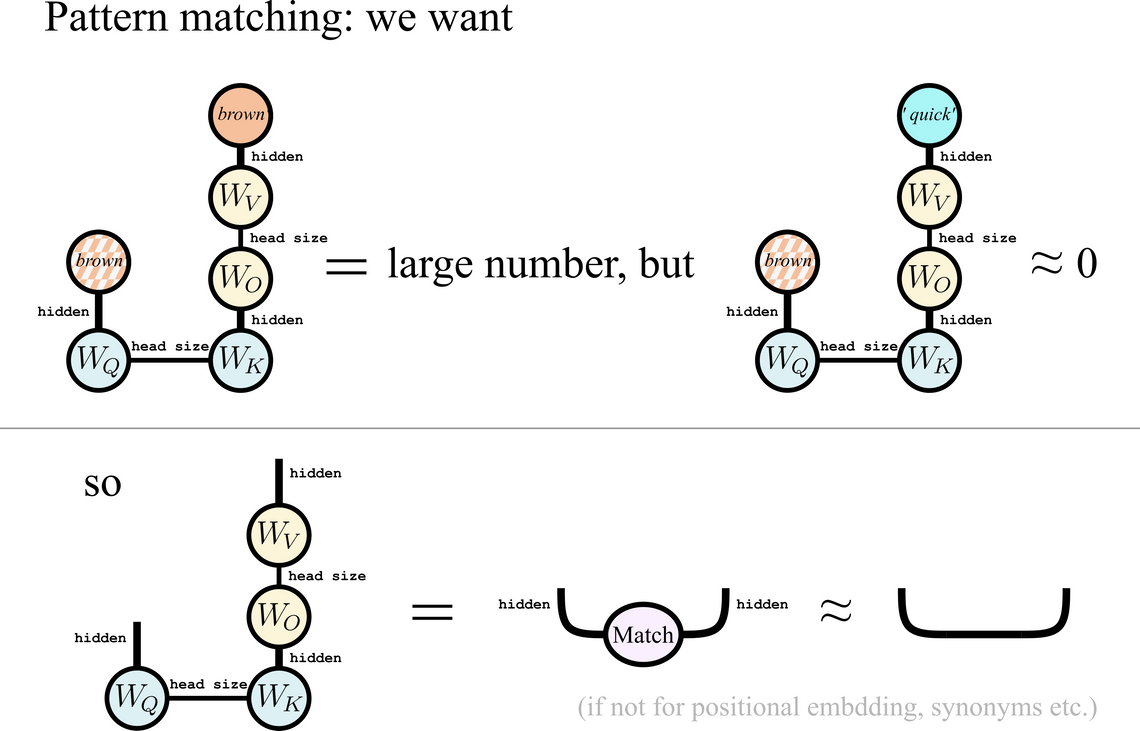}\;.
\]
So $W_O $ and $W_V $ in layer 1 (shown in yellow) and $W_Q $ and $W_K $ in layer 2 (shown in blue) can effectively just behave like identity matrices in the relevant subspace, or anything else producing something like a ``semantic delta tensor'' when you compose them together. We denote this tensor as ``Match'' because it should be near zero when contracted with any two vectors unless the vector on the left semantically ``matches'' with the vector on the right. 

Now, we don't really care about the matching token per se, we just want to know which token came after it. We can see that the ``key'' side of the attention pattern $A $ from layer 1 is going to be indexed at whatever token positions we get a match on (and multiplied by a number depending on strong the match is):
\[
\includegraphics[width=0.93\linewidth]{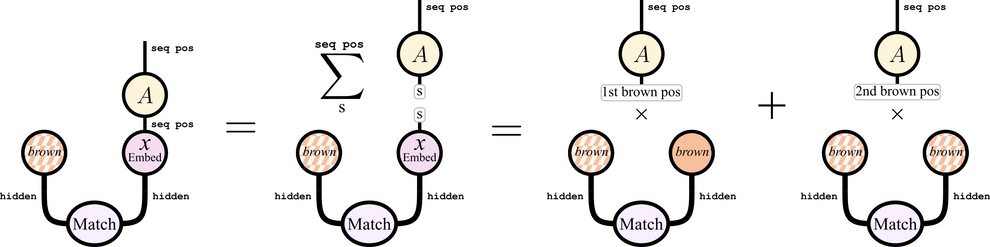}\;.
\]
We know that the token we want is in the position directly after  \code{1st brown pos}, so $A $ should map an index of \code{pos} on its key side to \code{pos + 1} on its query side so that we can index that token. Then we'll have
\[
\includegraphics[width=0.27\linewidth]{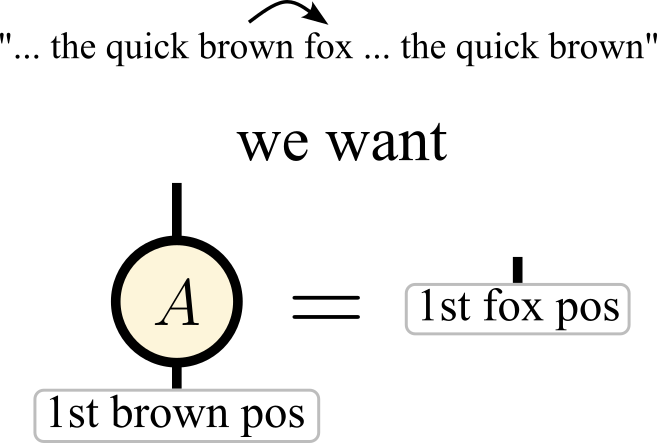}\;,
\]
as we desire. We see that an attention pattern $A $ doing this can just be a fixed off-diagonal delta-tensor $A_{k,\,q}=\delta_{k,\,q+1} $: 
\[
\includegraphics[width=0.67\linewidth]{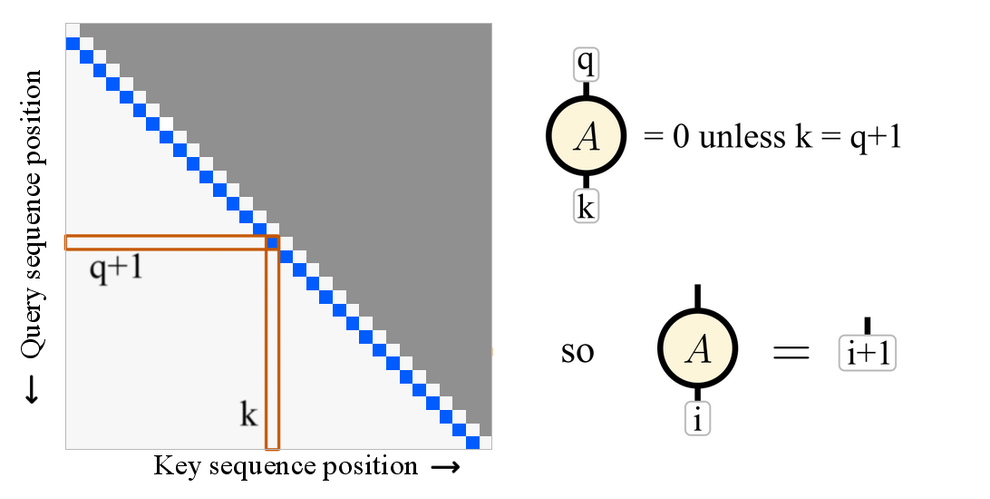}\;.
\]
This is also equivalent to mapping \code{pos} on its query side to \code{pos - 1} on its key side: always just attending to the previous token, so heads with this attention pattern are known as ``previous token heads''. This attention pattern also removes the unwanted \code{2nd brown pos} term, because this term indexes the final column of the attention matrix where all entries are zero. 

Putting this attention pattern $A_{k,\,q}=\delta_{k,\,q+1} $ into our circuit therefore simplifies it to:
\[
\includegraphics[width=0.97\linewidth]{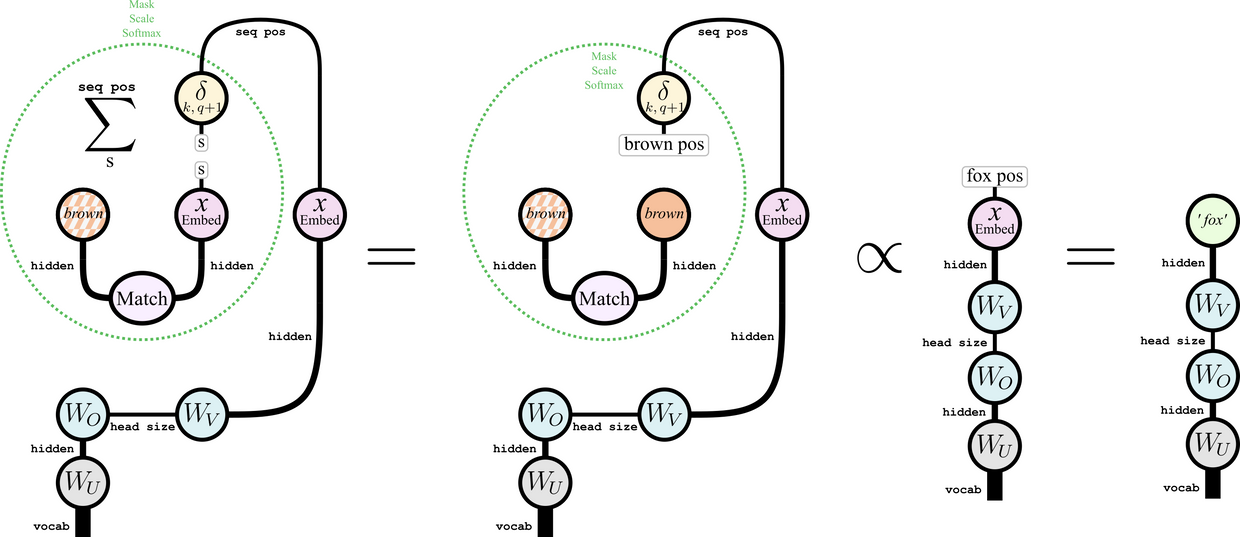}\;,
\]
so we can get the desired ``fox'' token out, so long as it gets properly handled by $W_{V} $ and $W_O $ (which are in charge of putting the ``fox'' information into the residual stream), and by the unembedding $W_U $ (which is in charge of getting the ``fox'' information out of the residual stream and turning it into the correct ``fox'' token). 

Putting everything together, we see that it's possible for virtual induction heads to have a very simple approximate form, composed almost entirely of delta-like tensors:
\[
\includegraphics[width=0.97\linewidth]{66-composition_to_fox.png}\;.
\]

We can sanity check this toy induction head by computing its attention pattern numerically on a repeating sequence of random vectors. We see that it looks like a real induction-head pattern, attending to the tokens which followed the current token previously in the sequence:
\[
\includegraphics[width=0.93\linewidth]{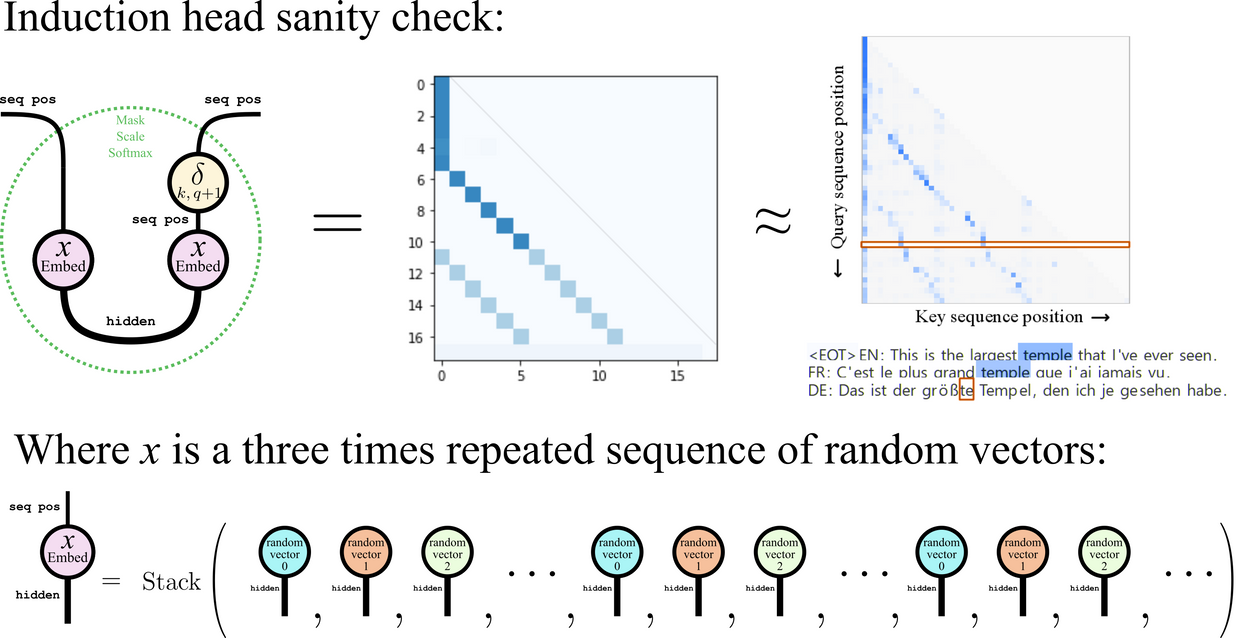}\;,
\]
here's the code for that:
\begin{lstlisting}[language=Python]
import torch as t
import matplotlib.pyplot as plt 
from einops import einsum, repeat 
hidden_dim = 768 
pattern_len = 6 
# Generate a three times repeated sequence of 6 random vectors 
x = t.rand((pattern_len, hidden_dim)) 
x = repeat(x, 'seq hidden -> (repeat seq) hidden', repeat=3) 
seq_len = x.shape[0] 
# Calculate the toy attention pattern 
prev_token_head = t.diag(t.tensor([1.0]*(seq_len-1)), diagonal=-1) 
match = t.eye(hidden_dim) 
attn_pattern = einsum(x, match, x, prev_token_head, 'seq0 hidden0, hidden0 hidden1, seq1 hidden1, seq1 seq2 -> seq2 seq0') 
# Apply the mask and softmax to the attention pattern 
attn_pattern = (t.tril(attn_pattern)-t.triu(t.ones_like(attn_pattern)*1e5)) 
attn_pattern = attn_pattern.softmax(dim=-1) 
print('attn_pattern = ')
plt.imshow(abs(attn_pattern), cmap='Blues'); plt.show()
\end{lstlisting}

However this virtual induction head formed by K-composition is just one term in the path expansion, so we'd need to make sure that this is actually the dominant term by suppressing the others. This is where the $W_{Q/K/O/V/U} $ matrices are important, because they selectively determine what gets taken from and added into the residual stream, allowing them to suppress unwanted terms in the path expansion (if they so choose) so that only this virtual induction head is important.  

Finally, we can put these $W_{O/V/U} $ matrices back in and see the full induction network rather than just the ``virtual attention head'' term in the path expansion:
\[
\includegraphics[width=0.43\linewidth]{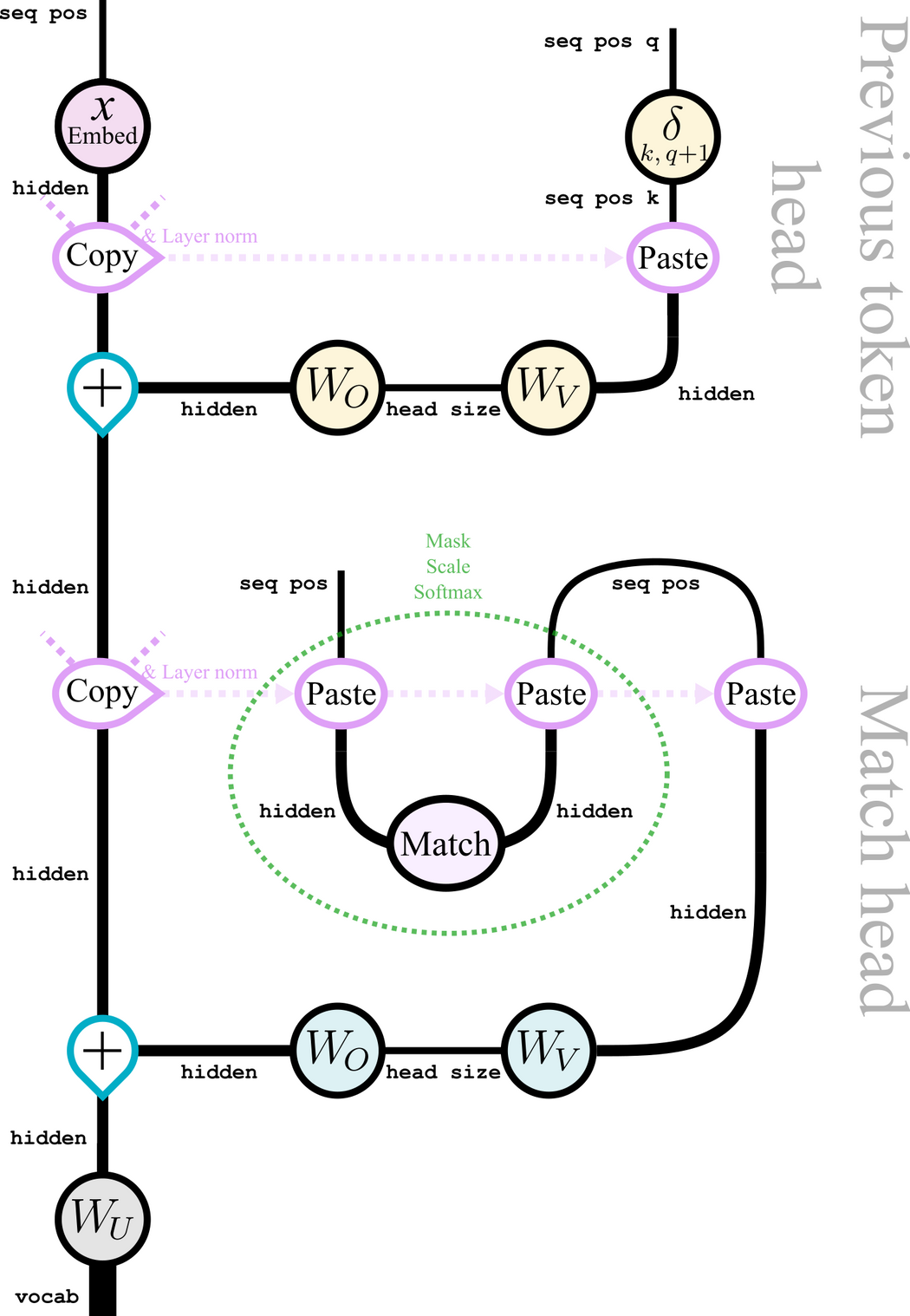}\;.
\]

\section{Conclusion}

Overall graphical tensor notation is a useful tool for understanding and communicating interpretability results, especially for anything involving operations between more than a few tensors. It surfaces dualities and interesting equivalences more easily than other notation, and remains intuitive without necessarily losing any mathematical rigor. It may just be my preference, but I frequently run into papers where I wish this notation was used. 

\section*{Acknowledgements}
I thank my PhD advisor Ian McCulloch for introducing me to tensor networks and the associated graphical notation, and for much support in working with these things over the years. Thanks also to the attendees and mentors at the 2nd \href{https://www.redwoodresearch.org/mlab}{Machine Learning for Alignment Bootcamp (MLAB)}, and to authors of excellent existing explanations of graphical tensor notation, such as in \href{https://www.math3ma.com/blog/matrices-as-tensor-network-diagrams}{the math3ma blog}, \href{https://simonverret.github.io/2019/02/16/tensor-network-diagrams-of-typical-neural-network.html}{Simon Verret's blog}, \href{https://tensornetwork.org/diagrams/}{tensornetwork.org}, \href{https://www.tensors.net/p-tutorial-1}{tensors.net}, and \href{https://arxiv.org/abs/1603.03039}{Hand-waving and Interpretive Dance: An Introductory Course on Tensor Networks}.  This work was supported by an Australian Government Research Training Program (RTP) Scholarship, and by \href{https://www.aisafetysupport.org/}{AI Safety Support} through the \href{https://www.matsprogram.org/}{ML Alignment \& Theory Scholars Program (MATS)}.

\vspace{1cm}
\textit{Editable SVG files for all diagrams in this document are available at \url{https://drive.google.com/drive/folders/16uuvkG1NtjhpAbchJ2R1Zp96IoZYKtk5}.}
 
\bibliographystyle{unsrtnat}
\bibliography{tensor_notation_interp}

\end{document}